\documentclass[sigconf,nonacm]{acmart}
\usepackage{adjustbox}
\usepackage{algorithm}
\usepackage{algpseudocode}
\usepackage{fontawesome5}
\usepackage{multirow}
\usepackage{pgfplots}
\usepackage{soul}
\usepackage{tikz}
\usepackage{xcolor}
\usepackage{cleveref}

\pgfplotsset{compat=newest,
    every axis/.append style={
        axis line style={draw=none},
        tick style={draw=none},
        xticklabel=\empty,
        yticklabel=\empty
    }
}

\usetikzlibrary{
    arrows.meta,
    backgrounds,
    calc,
    positioning,
    shadows,
    shapes.misc,
    intersections,
}

\newcommand{\distplot}[3]{
    \begin{tikzpicture}
        \begin{axis}[
            xmin=-3, xmax=3,
            ymin=-0.1, ymax=.9,
        ]
            \addplot[no marks, smooth, white, line width=30pt, samples=100]    {(2*pi*#2^2)^(-1/2)*e^(-(x-#1)^2/(2*#2^2))};
            \addplot[no marks, smooth, #3, line width=10pt, samples=100]    {(2*pi*#2^2)^(-1/2)*e^(-(x-#1)^2/(2*#2^2))};
        \end{axis}
    \end{tikzpicture}
}

\definecolor{hgfblue}{RGB}{0, 90, 160}
\colorlet{hgfblue10}{hgfblue!10!white}
\colorlet{hgfblue20}{hgfblue!20!white}
\colorlet{hgfblue30}{hgfblue!30!white}
\colorlet{hgfblue40}{hgfblue!40!white}
\colorlet{hgfblue50}{hgfblue!50!white}
\colorlet{hgfblue60}{hgfblue!60!white}
\colorlet{hgfblue70}{hgfblue!70!white}
\colorlet{hgfblue80}{hgfblue!80!white}
\colorlet{hgfblue90}{hgfblue!90!white}

\definecolor{hgfdarkblue}{RGB}{0, 40, 100}
\colorlet{hgfdarkblue10}{hgfdarkblue!10!white}
\colorlet{hgfdarkblue20}{hgfdarkblue!20!white}
\colorlet{hgfdarkblue30}{hgfdarkblue!30!white}
\colorlet{hgfdarkblue40}{hgfdarkblue!40!white}
\colorlet{hgfdarkblue50}{hgfdarkblue!50!white}
\colorlet{hgfdarkblue60}{hgfdarkblue!60!white}
\colorlet{hgfdarkblue70}{hgfdarkblue!70!white}
\colorlet{hgfdarkblue80}{hgfdarkblue!80!white}
\colorlet{hgfdarkblue90}{hgfdarkblue!90!white}

\definecolor{hgflightblue}{RGB}{20, 200, 255}
\colorlet{hgflightblue10}{hgflightblue!10!white}
\colorlet{hgflightblue20}{hgflightblue!20!white}
\colorlet{hgflightblue30}{hgflightblue!30!white}
\colorlet{hgflightblue40}{hgflightblue!40!white}
\colorlet{hgflightblue50}{hgflightblue!50!white}
\colorlet{hgflightblue60}{hgflightblue!60!white}
\colorlet{hgflightblue70}{hgflightblue!70!white}
\colorlet{hgflightblue80}{hgflightblue!80!white}
\colorlet{hgflightblue90}{hgflightblue!90!white}

\definecolor{hgfgreen}{RGB}{140, 180, 35}
\colorlet{hgfgreen10}{hgfgreen!10!white}
\colorlet{hgfgreen20}{hgfgreen!20!white}
\colorlet{hgfgreen30}{hgfgreen!30!white}
\colorlet{hgfgreen40}{hgfgreen!40!white}
\colorlet{hgfgreen50}{hgfgreen!50!white}
\colorlet{hgfgreen60}{hgfgreen!60!white}
\colorlet{hgfgreen70}{hgfgreen!70!white}
\colorlet{hgfgreen80}{hgfgreen!80!white}
\colorlet{hgfgreen90}{hgfgreen!90!white}

\definecolor{hgfgray}{RGB}{90, 105, 110}
\colorlet{hgfgray10}{hgfgray!10!white}
\colorlet{hgfgray20}{hgfgray!20!white}
\colorlet{hgfgray30}{hgfgray!30!white}
\colorlet{hgfgray40}{hgfgray!40!white}
\colorlet{hgfgray50}{hgfgray!50!white}
\colorlet{hgfgray60}{hgfgray!60!white}
\colorlet{hgfgray70}{hgfgray!70!white}
\colorlet{hgfgray80}{hgfgray!80!white}
\colorlet{hgfgray90}{hgfgray!90!white}

\definecolor{hgfhighlight}{RGB}{205, 238, 251}
\colorlet{hgfhighlight10}{hgfhighlight!10!white}
\colorlet{hgfhighlight20}{hgfhighlight!20!white}
\colorlet{hgfhighlight30}{hgfhighlight!30!white}
\colorlet{hgfhighlight40}{hgfhighlight!40!white}
\colorlet{hgfhighlight50}{hgfhighlight!50!white}
\colorlet{hgfhighlight60}{hgfhighlight!60!white}
\colorlet{hgfhighlight70}{hgfhighlight!70!white}
\colorlet{hgfhighlight80}{hgfhighlight!80!white}
\colorlet{hgfhighlight90}{hgfhighlight!90!white}

\definecolor{hgfmint}{RGB}{5, 229, 186}
\colorlet{hgfmint10}{hgfmint!10!white}
\colorlet{hgfmint20}{hgfmint!20!white}
\colorlet{hgfmint30}{hgfmint!30!white}
\colorlet{hgfmint40}{hgfmint!40!white}
\colorlet{hgfmint50}{hgfmint!50!white}
\colorlet{hgfmint60}{hgfmint!60!white}
\colorlet{hgfmint70}{hgfmint!70!white}
\colorlet{hgfmint80}{hgfmint!80!white}
\colorlet{hgfmint90}{hgfmint!90!white}

\definecolor{hgfpale}{RGB}{236, 251, 253}
\colorlet{hgfpale10}{hgfpale!10!white}
\colorlet{hgfpale20}{hgfpale!20!white}
\colorlet{hgfpale30}{hgfpale!30!white}
\colorlet{hgfpale40}{hgfpale!40!white}
\colorlet{hgfpale50}{hgfpale!50!white}
\colorlet{hgfpale60}{hgfpale!60!white}
\colorlet{hgfpale70}{hgfpale!70!white}
\colorlet{hgfpale80}{hgfpale!80!white}
\colorlet{hgfpale90}{hgfpale!90!white}

\definecolor{hgfaerospace}{RGB}{80, 200, 170}
\definecolor{hgfast}{named}{hgfaerospace}

\colorlet{hgfaerospace10}{hgfaerospace!10!white}
\colorlet{hgfaerospace20}{hgfaerospace!20!white}
\colorlet{hgfaerospace30}{hgfaerospace!30!white}
\colorlet{hgfaerospace40}{hgfaerospace!40!white}
\colorlet{hgfaerospace50}{hgfaerospace!50!white}
\colorlet{hgfaerospace60}{hgfaerospace!60!white}
\colorlet{hgfaerospace70}{hgfaerospace!70!white}
\colorlet{hgfaerospace80}{hgfaerospace!80!white}
\colorlet{hgfaerospace90}{hgfaerospace!90!white}

\colorlet{hgfast10}{hgfast!10!white}
\colorlet{hgfast20}{hgfast!20!white}
\colorlet{hgfast30}{hgfast!30!white}
\colorlet{hgfast40}{hgfast!40!white}
\colorlet{hgfast50}{hgfast!50!white}
\colorlet{hgfast60}{hgfast!60!white}
\colorlet{hgfast70}{hgfast!70!white}
\colorlet{hgfast80}{hgfast!80!white}
\colorlet{hgfast90}{hgfast!90!white}

\definecolor{hgfearthandenvironment}{RGB}{50, 100, 105}
\definecolor{hgfee}{named}{hgfearthandenvironment}

\colorlet{hgfearthandenvironment10}{hgfearthandenvironment!10!white}
\colorlet{hgfearthandenvironment20}{hgfearthandenvironment!20!white}
\colorlet{hgfearthandenvironment30}{hgfearthandenvironment!30!white}
\colorlet{hgfearthandenvironment40}{hgfearthandenvironment!40!white}
\colorlet{hgfearthandenvironment50}{hgfearthandenvironment!50!white}
\colorlet{hgfearthandenvironment60}{hgfearthandenvironment!60!white}
\colorlet{hgfearthandenvironment70}{hgfearthandenvironment!70!white}
\colorlet{hgfearthandenvironment80}{hgfearthandenvironment!80!white}
\colorlet{hgfearthandenvironment90}{hgfearthandenvironment!90!white}

\colorlet{hgfee10}{hgfee!10!white}
\colorlet{hgfee20}{hgfee!20!white}
\colorlet{hgfee30}{hgfee!30!white}
\colorlet{hgfee40}{hgfee!40!white}
\colorlet{hgfee50}{hgfee!50!white}
\colorlet{hgfee60}{hgfee!60!white}
\colorlet{hgfee70}{hgfee!70!white}
\colorlet{hgfee80}{hgfee!80!white}
\colorlet{hgfee90}{hgfee!90!white}

\definecolor{hgfenergy}{RGB}{255, 210, 40}

\colorlet{hgfenergy10}{hgfenergy!10!white}
\colorlet{hgfenergy20}{hgfenergy!20!white}
\colorlet{hgfenergy30}{hgfenergy!30!white}
\colorlet{hgfenergy40}{hgfenergy!40!white}
\colorlet{hgfenergy50}{hgfenergy!50!white}
\colorlet{hgfenergy60}{hgfenergy!60!white}
\colorlet{hgfenergy70}{hgfenergy!70!white}
\colorlet{hgfenergy80}{hgfenergy!80!white}
\colorlet{hgfenergy90}{hgfenergy!90!white}

\definecolor{hgfhealth}{RGB}{210, 50, 100}

\colorlet{hgfhealth10}{hgfhealth!10!white}
\colorlet{hgfhealth20}{hgfhealth!20!white}
\colorlet{hgfhealth30}{hgfhealth!30!white}
\colorlet{hgfhealth40}{hgfhealth!40!white}
\colorlet{hgfhealth50}{hgfhealth!50!white}
\colorlet{hgfhealth60}{hgfhealth!60!white}
\colorlet{hgfhealth70}{hgfhealth!70!white}
\colorlet{hgfhealth80}{hgfhealth!80!white}
\colorlet{hgfhealth90}{hgfhealth!90!white}

\definecolor{hgfinformation}{RGB}{160, 35, 90}
\definecolor{hginfo}{named}{hgfinformation}

\colorlet{hgfinformation10}{hgfinformation!10!white}
\colorlet{hgfinformation20}{hgfinformation!20!white}
\colorlet{hgfinformation30}{hgfinformation!30!white}
\colorlet{hgfinformation40}{hgfinformation!40!white}
\colorlet{hgfinformation50}{hgfinformation!50!white}
\colorlet{hgfinformation60}{hgfinformation!60!white}
\colorlet{hgfinformation70}{hgfinformation!70!white}
\colorlet{hgfinformation80}{hgfinformation!80!white}
\colorlet{hgfinformation90}{hgfinformation!90!white}

\colorlet{hgfinfo10}{hgfinformation!10!white}
\colorlet{hgfinfo20}{hgfinformation!20!white}
\colorlet{hgfinfo30}{hgfinformation!30!white}
\colorlet{hgfinfo40}{hgfinformation!40!white}
\colorlet{hgfinfo50}{hgfinformation!50!white}
\colorlet{hgfinfo60}{hgfinformation!60!white}
\colorlet{hgfinfo70}{hgfinformation!70!white}
\colorlet{hgfinfo80}{hgfinformation!80!white}
\colorlet{hgfinfo90}{hgfinformation!90!white}

\definecolor{hgfmatter}{RGB}{240, 120, 30}

\colorlet{hgfmatter10}{hgfmatter!10!white}
\colorlet{hgfmatter20}{hgfmatter!20!white}
\colorlet{hgfmatter30}{hgfmatter!30!white}
\colorlet{hgfmatter40}{hgfmatter!40!white}
\colorlet{hgfmatter50}{hgfmatter!50!white}
\colorlet{hgfmatter60}{hgfmatter!60!white}
\colorlet{hgfmatter70}{hgfmatter!70!white}
\colorlet{hgfmatter80}{hgfmatter!80!white}
\colorlet{hgfmatter90}{hgfmatter!90!white}

\AtBeginDocument{%
  }

\setcopyright{acmlicensed}
\copyrightyear{2026}
\acmYear{2026}
\acmDOI{XXXXXXX.XXXXXXX}
\acmConference[PASC26]{Platform for Advanced Scientific Computing}{June 29--July 01,
  2026}{Bern, CH}
\begin{document}

%%
%% The "title" command has an optional parameter,
%% allowing the author to define a "short title" to be used in page headers.
\title{Sampling Parallelism for Fast and Efficient Bayesian Learning}

%%
%% The "author" command and its associated commands are used to define
%% the authors and their affiliations.
%% Of note is the shared affiliation of the first two authors, and the
%% "authornote" and "authornotemark" commands
%% used to denote shared contribution to the research.
\author{Asena Karolin Özdemir}
\orcid{0009-0003-2808-4744}
%\orcid{1234-5678-9012}
%\author{G.K.M. Tobin}
%\authornotemark[1]
%\email{webmaster@marysville-ohio.com}
\affiliation{%
  \institution{Karlsruhe Institute of Technology}
  \city{Karlsruhe}
  \country{Germany}
}
\email{asena.oezdemir@kit.edu}

\author{Lars H. Heyen}
\orcid{0000-0001-7949-1858}
\affiliation{%
  \institution{Karlsruhe Institute of Technology}
  \city{Karlsruhe}
  \country{Germany}
}
\email{lars.heyen@kit.edu}

\author{Arvid Weyrauch}
\orcid{0000-0002-2684-0927}
\affiliation{%
  \institution{Karlsruhe Institute of Technology}
  \city{Karlsruhe}
  \country{Germany}
}
\email{arvid.weyrauch@kit.edu}

\author{Achim Streit}
\orcid{0000-0002-5065-469X}
\affiliation{%
  \institution{Karlsruhe Institute of Technology}
  \city{Karlsruhe}
  \country{Germany}
}
\email{achim.streit@kit.edu}

\author{Markus Götz}
\orcid{0000-0002-2233-1041}
\affiliation{%
  \institution{Helmholtz AI\\ Karlsruhe Institute of Technology}
  \city{Karlsruhe}
  \country{Germany}
}
\email{markus.goetz@kit.edu}

\author{Charlotte Debus}
\orcid{0000-0002-7156-2022}
\affiliation{%
  \institution{Karlsruhe Institute of Technology}
  \city{Karlsruhe}
  \country{Germany}
}
\email{charlotte.debus@kit.edu}

%%
%% By default, the full list of authors will be used in the page
%% headers. Often, this list is too long, and will overlap
%% other information printed in the page headers. This command allows
%% the author to define a more concise list
%% of authors' names for this purpose.
\renewcommand{\shortauthors}{Özdemir et al.}

%%
%% The abstract is a short summary of the work to be presented in the
%% article.
\begin{abstract}
%The application of machine learning methods, particularly deep neural networks, in risk-sensitive domains such as healthcare, environmental forecasting, and finance makes reliable quantification of uncertainty essential.

%But a lot of UQ methods are computationally expensive due to the memory and time costs of sampling. 

%To enable UQ research we present sampling parallelism, a method that parallelizes exactly what makes BNNs computationally expensive.

%We explain the methodology, demonstrate feasibility in weak and strong scaling experiment results as well as predictive performance results on an example task and architecture. Compare with DDP as a baseline and also demonstrate how it can work in conjunction with other parallelization techniques by doing a hybrid version.

%We show close to perfect weak scaling. While DDP beats sampling parallelism in strong scalin we show a benefit that loading the same data in different GPUs can have due to different augmentations and how it can lead  to faster convergence.

Machine learning models, and deep neural networks in particular, are increasingly deployed in risk-sensitive domains such as healthcare, environmental forecasting, and finance, where reliable quantification of predictive uncertainty is essential. However, many uncertainty quantification (UQ) methods remain difficult to apply due to their substantial computational cost. Sampling-based Bayesian learning approaches, such as Bayesian neural networks (BNNs), are particularly expensive since drawing and evaluating multiple parameter samples rapidly exhausts memory and compute resources. These constraints have limited the accessibility and exploration of Bayesian techniques thus far.
To address these challenges, we introduce sampling parallelism, a simple yet powerful parallelization strategy that targets the primary bottleneck of sampling-based Bayesian learning: the samples themselves. By distributing sample evaluations across multiple GPUs, our method reduces memory pressure and training time without requiring architectural changes or extensive hyperparameter tuning. We detail the methodology and evaluate its performance on a few example tasks and architectures, comparing against distributed data parallelism (DDP) as a baseline. We further demonstrate that sampling parallelism is complementary to existing strategies by implementing a hybrid approach that combines sample and data parallelism.
Our experiments show near-perfect scaling when the sample number is scaled proportionally to the computational resources, confirming that sample evaluations parallelize cleanly. Although DDP achieves better raw speedups under scaling with constant workload, sampling parallelism has a notable advantage: by applying independent stochastic augmentations to the same batch on each GPU, it increases augmentation diversity and thus reduces the number of epochs required for convergence.
%In strong-scaling settings, DDP achieves better raw speedups, but sampling parallelism offers a distinctive benefit: because each GPU receives the same batch but applies independent stochastic augmentations, models trained with sampling parallelism converge faster in terms of epochs due to the higher augmentation diversity. 
\end{abstract}



%%
%% Keywords. The author(s) should pick words that accurately describe
%% the work being presented. Separate the keywords with commas.
%\keywords{Do, Not, Use, This, Code, Put, the, Correct, Terms, for, Your, Paper}
%% A "teaser" image appears between the author and affiliation
%% information and the body of the document, and typically spans the
%% page.

%\received[accepted]{5 June 2009}

%%
%% This command processes the author and affiliation and title
%% information and builds the first part of the formatted document.
\maketitle
\setlength{\belowcaptionskip}{-5pt}
\section{Introduction}

Reliable uncertainty estimates for neural network predictions support the transparency, accuracy, and trustworthiness requirements that are anticipated to gain substantial importance under the EU AI Act~\cite{valdenegro2024dilemma}. As such, scalable and efficient methods for uncertainty quantification (UQ) are essential to provide such estimates,  especially in high-risk domains such as atmospheric forecasting ~\cite{lam2023learning}, healthcare~\cite{abdullah2022review}, and finance~\cite{mashrur2020machine}, where critical decisions directly depend on model outputs. 

Numerous approaches to quantify different aspects of uncertainty have been proposed, and recent trends have demonstrated the benefit of probabilistic loss functions or the utilization of diffusion models to quantify uncertainty~\cite{price2023gencast, lessig2023atmorep}. However, probabilistic treatment of a neural network itself, especially in a fully Bayesian framework, is still lacking, as the high computational cost associated with training these models remains a key barrier. Most existing methods rely on stochastic sampling from probability distributions, which is expensive both in time and memory. The sampling process not only slows down training but also requires storing multiple sampled model instances, often turning into a performance bottleneck, or even becoming infeasible when the number of required samples exceeds available GPU memory~\cite{andrade2025effectiveness} .

Despite the growing demand, practical implementation of state-of-the-art \emph{sampling-based Bayesian learning} in neural networks is oftentimes restricted to small proof-of-concept models and applications~\cite{gawlikowski2023survey}. 

However, modern networks now routinely contain billions of parameters across domains such as molecular modeling~\cite{krishna2024generalized, jumper2021highly}, climate and weather prediction~\cite{bi2023accurate,  kurth2023fourcastnet}, and large-scale language modeling~\cite{brown2020language, hoffmann2022training}, making sampling-based Bayesian learning computationally infeasible.
Research on practical and scalable approaches for sampling-based Bayesian learning remains limited, even though there are many use-cases in which having uncertainty estimates would be beneficial.

To address these challenges, we explore sampling parallelism, a strategy designed to improve runtime and memory efficiency in uncertainty-aware neural network training using sampling-based Bayesian learning. By reducing the computational barriers associated with sampling, sampling parallelism enables broader experimentation and deployment of these methods. In an exemplary experimental evaluation, we demonstrate that sampling parallelism provides a complementary scaling axis to existing parallelization approaches for neural networks and can aid in pushing scalability limits while even improving model convergence.

\section{Background}
Uncertainty in neural networks can have two sources: the inherent randomness of the world (aleatoric, data driven), and the lack of knowledge (epistemic, model driven)~\cite{kendall2017uncertainties}, both of which need to be quantified to navigate risk-sensitive applications of machine learning. There are a variety of methods that enable UQ, however, they oftentimes require sampling which can be computationally expensive as well as memory intensive~\cite{lakshminarayanan2017simple, neal2012bayesian, maddox2019simple, gal2016dropout}. In this paper, we discuss Bayesian neural networks (BNNs) with mean-field variational inference, and Monte Carlo dropout (MCD) as examples, however the introduced parallelization concept can be applied to other sampling-based methods such as ensemble models ~\cite{lakshminarayanan2017simple}, Markov Chain Monte Carlo (MCMC) algorithms ~\cite{chowdhury2018parallel} or sampling in General Adversarial Networks (GANs) ~\cite{goodfellow2020generative} without loss of generality.

\subsection{Bayesian Neural Networks and Variational Inference}
In BNNs the goal is to learn the probability distribution over all possible outputs, given the inputs and the training data. Mathematically, a BNN is expressed as
\[
p(y^{*} \mid x^{*}, \mathcal{D})
    = \int p(y^{*} \mid x^{*}, \mathbf{w}) \, 
      p(\mathbf{w} \mid \mathcal{D}) \, d\mathbf{w}
\]
where $\mathcal{D}$ is the training data, $\mathbf{w}$ are the model weights, $x^*$ is the input and $y^*$ is the output.

%However $p(\mathbf{w} \mid \mathcal{D})$, which is called the posterior, can be high dimensional and non-gaussian, which makes the computation of this term intractable. 
However, the computation of the so-called \emph{posterior} $p(\mathbf{w} \mid \mathcal{D})$ is generally intractable.
To circumvent this, the method of \emph{Variational Inference (VI)}~\cite{graves2011practical, blundell2015weight} approximates the posterior by optimizing a parametrized distribution $q(\mathbf{w} \mid \theta)$ such that $q(\mathbf{w} \mid \theta) \approx p(\mathbf{w} \mid \mathcal{D})$.
The most common choice for $q(\mathbf{w} \mid \theta)$ is a Gaussian distribution where the parameters $\theta$ are the mean $\mu$ and the standard deviation $\sigma$, i.e. each model weight is described by a $\mu$ and $\sigma$ value. 

With this approximation, the loss function of the optimization problem in the Bayesian model formulation above can be reformulated to maximizing the so-called Evidence Lower Bound (ELBO):
\[
\mathcal{L}(\theta) 
= \mathbb{E}_{q(\mathbf{w} \mid \theta)} 
  \left[ \log p(\mathcal{D} \mid \mathbf{w}) \right]
  - \mathrm{KL}\!\left( q(\mathbf{w} \mid \theta) 
      \,\|\, p(\mathbf{w}) \right)
\]
where the first term $\mathbb{E}_{q(\mathbf{w} \mid \theta)} 
\left[ \log p(\mathcal{D} \mid \mathbf{w}) \right]$ is the data fitting term, which describes how well the current posterior approximation $q(\mathbf{w} \mid \theta)$ fits the data. The second term $-\mathrm{KL}\!\left( q(\mathbf{w} \mid \theta) \,\|\, p(\mathbf{w}) \right)$ is the prior matching term. $p(\mathbf{w})$ represents the assumptions that are made about the posterior before any data has been taken into account. Since this distribution captures prior assumptions, it is referred to as the \emph{prior}. The prior matching term draws the current posterior $q(\mathbf{w} \mid \theta)$ towards the prior, by minimizing the Kullback-Leibler (KL) divergence, which is a measure that indicates distinguishability between two probability distributions.

Practically, training a BNN with VI is performed as follows: First, the parameter values $\theta$ of the prior $q(\mathbf{w} \mid \theta)$ are initialized. Choosing a Gaussian prior, the means of the weight distribution are initialized in the same fashion as for a non-Bayesian network, while the standard deviations are initialized to a constant value that depends on the layer size. For each epoch and each batch, $n$ sets of parameters (random samples) are drawn from the current weight distributions. For each of these sampled model weights, a forward pass is performed to obtain a model prediction. The distribution over these predictions from all sampled weights are aggregated to obtain an averaged prediction as well as an uncertainty in terms of a standard-deviation of the prediction. The ELBO (rescaled with the dataset size to avoid overflow) is calculated and the backwards pass is performed to update the parameters of the weight distributions. \Cref{bnn-train-regular} illustrates the algorithmic flow of the described procedure.

\begin{algorithm}%[H]
\caption{Bayesian Neural Network Training with Variational Inference \label{bnn-train-regular}}
\begin{algorithmic}[1]
\Require Dataset $\mathcal{D}$, number of epochs $E$, number of samples $S$
\State Initialize variational parameters $\mu$ and $\sigma$ for each weight % $\sigma = \text{softplus}(\rho)$
\For{$\text{epoch} = 1$ to $E$}
    \For{each minibatch $(x, y) \in \mathcal{D}$}
        \For {$\text{s} = 1$ to $S$}
        \State Sample $\epsilon_s \sim \mathcal{N}(0, I)$
        \State $w_s \gets \mu + \sigma \odot \epsilon_s$ \hfill // reparameterization trick
        \State $\hat{y}_s \gets \text{ForwardPass}(x, w_s)$
        \EndFor
        \State $\mathcal{L}_{\text{data}} \gets \frac{1}{S} \sum_{s=1}^{S} \text{Loss}(\hat{y}_s, y)$
        \State $\mathcal{L}_{\text{KL}} \gets \frac{1}{2} \sum_i \big(\sigma_i^2 + \mu_i^2 - 1 - \log \sigma_i^2 \big)$ // Prior Matching Term
        \State $\mathcal{L}_s \gets \mathcal{L}_{\text{data}} + \frac{1}{|\mathcal{D}|} \mathcal{L}_{\text{KL}}$ \hfill // negative scaled ELBO
        \State Compute gradients $\nabla \mathcal{L}(\mu, \sigma)$
        \State Update $\mu$ and $\sigma$ using optimizer (e.g., Adam)
    \EndFor
\EndFor
\end{algorithmic}
\end{algorithm}

\subsection{Monte Carlo Dropout}
BNNs are very powerful tools for UQ; however, they tend to be difficult to implement and train to sufficient accuracy, especially given the high computational demand and memory footprint. A simplified approach that is often used is MCD, which can be interpreted as a variational approximation to the posterior over the network weights, effectively multiplying each weight with a Bernoulli distribution.
%In this approach dropout, which is normally a regularization method to prevent underfitting, is used in a different context. This way, variational inference can be approximated without the complexities of BNNs.

Normally, dropout is used as a regularization technique to prevent overfitting in neural networks~\cite{JMLR:v15:srivastava14a}. It randomly sets a subset of neurons to zero during training, effectively training an ensemble of sub-networks. At inference time, dropout is disabled to use the model at full capacity.
\citet{gal2016dropout} demonstrated that enabling dropout at inference time leads to a form of approximate Bayesian inference. Multiple forward passes with different dropout masks produce a distribution of predictions, where the mean serves as the model’s output and the standard deviation captures epistemic uncertainty. The number of forward passes, $S$, determines the fidelity of the predictive distribution: larger $S$ give more accurate uncertainty estimates but increases inference time. 

While MCD does not lead to a full Bayesian neural network, it offers a practical and easy-to-implement approximation without modifying the model architecture. However, the requirement of performing multiple forward passes for a single prediction slows down inference, particularly for large models. Thus, MCD suffers from similar problems as full BNNs regarding computational complexity and memory demands of samples.

\section{Related Works}
The general approach to deal with time and memory constraints in computational problems is to parallelize and distribute the problem to multiple processes. For neural network training, parallelization can be performed along different axes. 

\subsection{Distributed Data Parallelism}
In many cases, the primary obstacle to fast training is the need to process large, complex datasets efficiently. Modern neural network training often involves iterating over millions of data points in a sequential batched manner, leading to a time-consuming process. As a result, a practical strategy for accelerating training is to distribute the work associated with each batch, including data loading as well as the forward and backward passes, across multiple GPUs. By partitioning the dataset into distinct shards and assigning them to different compute units, the training of separate portions of the data can proceed in parallel, reducing overall training time and improving hardware utilization.

This approach is referred to as Distributed Data Parallelism (DDP) and is a widely used strategy for scaling the training of deep neural networks across multiple accelerator devices, i.e., GPUs. Since only the data is parallelized in DDP, each GPU still holds a full replica of the model. After each GPU loads its share of the data on which it performs the forward and backward pass, the locally calculated gradients are synchronized across all GPUs in an allreduce operation to update the model instances on all GPUs. This approach allows for balanced workload distribution, and advanced strategies to efficiently overlap computation and communication, leading to near-linear speedup in most cases. The existence of optimized off-the-shelf frameworks such as torch-distributed~\cite{paszke2019pytorch, li2020pytorch} has led to a wide acceptance and implementation of the method. Even though communication can become a bottleneck at scale, DDP is still considered the most effective method to speed up training.  

One issue that arises when using DDP at scale is that the effective batch size increases linearly with the number of GPUs, since each GPU handles a mini-batch independently.
While larger batch sizes can stabilize training, they can also reduce overall model accuracy and generalization due to convergence to sharper minima~\cite{goyal2017accurate, keskar2016large}. This effect is known as \emph{large batch effects}. To circumvent large batch effects and maintain convergence, the learning rate and optimization schedule can be adjusted. However, such tuning is often based on heuristics, that are not always effective in every application.

Another challenge in DDP is the fact that the entire model, including all parameters, intermediate activations, and optimizer states, needs to fit into the memory of a single GPU. This results in a non-negligible memory demand, which becomes even more pressing in the Bayesian setting, where multiple samples of the model are drawn from the parameter distribution. While DDP provides the advantage that more batches can be processed in parallel, it becomes thus less suitable for scalable sampling-based Bayesian learning, when multiple samples of very large models need to be stored in GPU memory.

\subsection{Model Parallelism}
With growing model sizes, the memory demand associated with holding all model parameters in GPU memory becomes increasingly challenging. In response, model parallelism (MP) has emerged as a way to distribute model parameters across GPUs and, in some cases, even parallelize the computation. Different ways to split the model exist, with each of them coming with its own benefits as well as caveats ~\cite{brakel2024model}. 

In \emph{tensor parallelism}, each layer within a neural network is split across multiple GPUs, and the mathematical operations within each layer are performed in parallel. This enables the model to train even when individual layers are too large to fit into memory. Tensor parallelism can be effective when communication between GPUs is quite fast. However, given the complex operations and layer types that are used in modern neural networks, tensor parallelism presents with substantial communication overheads, and requires careful orchestration and dedicated fine-tuning on very fast interconnects to be efficient. 
A simpler alternative is \emph{pipeline parallelism}, in which networks are distributed across layers, i.e., individual layers or groups thereof are distributed across GPUs~\cite{narayanan2019pipedream, huang2019gpipe}. However, while pipeline parallelism is easy to implement and can efficiently address memory bottlenecks for large models, the sequential nature of the forward-backward-pass computation prohibits true parallel execution.

As neural network architectures advance, so do the strategies to distribute their computational load. In particular, the rise of Transformer-based architectures has pushed the field, given the massive size of these models and the memory demand associated with the attention mechanism. The most common approach among them is MegatronLM~\cite{shoeybi2019megatron}, a parallelization strategy specifically taylored to Transformer blocks, which has been proven to work efficiently and is available as off-the-shelf library for widespread use.

\subsection{Sharded Parallelism }
Some methods don't fit into either the data parallelism nor model parallelism category, because they share some properties with both. One of them is sharded parallelism, which distributes a model’s parameters, gradients, and optimizer states across GPUs. This greatly reduces memory usage while preserving the standard data-parallel compute pattern: each GPU still performs the full forward and backward pass on its own batch. Before executing each layer, the required parameter shards are temporarily gathered, used for computation, and then released so that only local shards remain. This makes sharded parallelism both data-parallel in terms of computation and model-parallel in terms of storage \cite{zhao2023pytorch}.

Compared to standard Distributed Data Parallel (DDP), which fully replicates the model on every device, sharded parallelism enables training larger models by avoiding redundant memory copies. Fully sharded Data Parallel (FSDP) implements this approach by performing layer-by-layer gather–compute–shard cycles. Although this increases communication relative to plain DDP, it provides substantial memory efficiency and scalability benefits for large models.

\subsection{Parallelization of Uncertainty Quantification and Sampling-Based Methods}
While other parallelization techniques that we discussed previously can be applied to uncertainty quantification methods as well, there are also parallelization approaches that specifically target UQ methods. Deep ensembles are inherently parallelizable, as independent models can be trained concurrently across multiple devices without the need for communication in between training steps, yielding both predictive performance and well-calibrated uncertainty estimates ~\cite{lakshminarayanan2017simple}.
Diffusion models use sampling in their denoising steps~\cite{ho2020denoising}, which are usually sequential and time-consuming since many of them are required just to process a single data item. Recent work demonstrates that these sequential steps can be parallelized for better performance ~\cite{shih2023parallel}. However, to the best of our knowledge, sampling parallelism in the context of BNNs and MCD has not been explored so far. While it may seem trivially parallel at first, there are several potential pitfalls, for example, the exact vs. approximate communication of the sample averages, which we elaborate on below. Moreover, the combination with data augmentation towards faster model convergence marks another innovation of our approach, which so far has not been explored. 

\section{Methodology}
Sampling-based UQ methods, and BNNs in particular, are inherently expensive to train, and pose substantial challenges with respect to computational and memory demand.
For one, they comprise more parameters than their deterministic counterparts, since each weight is replaced by a distribution that is characterized by at least two parameters instead of one, making them by a factor larger.
Moreover, drawing $s$ parameter samples during a forward pass effectively scales the model size by a factor of $s$, further augmenting the memory footprint. In addition, generating these samples and performing separate forward and backward passes for each of them substantially increases computational cost.

These factors have hindered the implementation of large-scale UQ methods in practice thus far. The goal of our work is to target these challenges through parallelization and by that improve the accessibility of research on BNNs and related UQ methodologies beyond toy examples.

\subsection{Sampling Parallelism}
\begin{figure}[ht]
    \centering
    \newcommand{\model}[2]{
    
    \pgfmathsetmacro{\xoffset}{#2 * 2.2}
    \pgfmathsetmacro{\yoffset}{#2 * 2.0}

    \node[draw=black, circle, thick, inner sep=0, minimum width=#2 cm] (n11) at (#1) {};

    \node[draw=black, circle, thick, inner sep=0, minimum width=#2 cm, below left=\yoffset and \xoffset of n11.south, anchor=center] (n21) {};
    \node[draw=black, circle, thick, inner sep=0, minimum width=#2 cm, below=\yoffset of n11.south, anchor=center] (n22) {};
    \node[draw=black, circle, thick, inner sep=0, minimum width=#2 cm, below right=\yoffset and \xoffset of n11.south, anchor=center] (n23) {};
    
    \node[draw=black, circle, thick, inner sep=0, minimum width=#2 cm, below=\yoffset of n21.south, anchor=center] (n31) {};
    \node[draw=black, circle, thick, inner sep=0, minimum width=#2 cm, below=\yoffset of n22.south, anchor=center] (n32) {};
    \node[draw=black, circle, thick, inner sep=0, minimum width=#2 cm, below=\yoffset of n23.south, anchor=center] (n33) {};
    
    \node[draw=black, circle, thick, inner sep=0, minimum width=#2 cm, below=\yoffset of n32.south, anchor=center] (n41) {};

    \draw[-{Stealth[scale=0.5]}] (n11) -- (n21);
    \draw[-{Stealth[scale=0.5]}] (n11) -- (n22);
    \draw[-{Stealth[scale=0.5]}] (n11) -- (n23);
    
    \draw[-{Stealth[scale=0.5]}] (n21) -- (n31);
    \draw[-{Stealth[scale=0.5]}] (n22) -- (n32);
    \draw[-{Stealth[scale=0.5]}] (n23) -- (n33);
    
    \draw[-{Stealth[scale=0.5]}] (n31) -- (n41);
    \draw[-{Stealth[scale=0.5]}] (n32) -- (n41);
    \draw[-{Stealth[scale=0.5]}] (n33) -- (n41);
}

\newcommand{\modelwithdists}[4]{
    \pgfmathsetmacro{\xoffset}{#2 * 2.2}
    \pgfmathsetmacro{\yoffset}{#2 * 2.0}

    \pgfmathsetmacro{\curvesize}{\xoffset * 12}
    \pgfmathsetmacro{\curveshift}{\yoffset * 3.5}

    \pgfmathdeclarerandomlist{means}{{-2}{-1.5}{-1}{-.5}{0}{1}{1.5}{2}}
    \pgfmathdeclarerandomlist{stds}{{.5}{.6}{.7}{.8}{.9}{1}}

    \node[draw=black, circle, thick, inner sep=0, minimum width=#2 cm] (n11) at (#1) {};

    \node[draw=black, circle, thick, inner sep=0, minimum width=#2 cm, below left=\yoffset and \xoffset of n11.south, anchor=center] (n21) {};
    \node[draw=black, circle, thick, inner sep=0, minimum width=#2 cm, below=\yoffset of n11.south, anchor=center] (n22) {};
    \node[draw=black, circle, thick, inner sep=0, minimum width=#2 cm, below right=\yoffset and \xoffset of n11.south, anchor=center] (n23) {};
    
    \node[draw=black, circle, thick, inner sep=0, minimum width=#2 cm, below=\yoffset of n21.south, anchor=center] (n31) {};
    \node[draw=black, circle, thick, inner sep=0, minimum width=#2 cm, below=\yoffset of n22.south, anchor=center] (n32) {};
    \node[draw=black, circle, thick, inner sep=0, minimum width=#2 cm, below=\yoffset of n23.south, anchor=center] (n33) {};
    
    \node[draw=black, circle, thick, fill=#3, inner sep=0, minimum width=#2 cm, below=\yoffset of n32.south, anchor=center] (n41) {};

    \foreach \i in {1, ..., 3}{
        \draw[-{Stealth[scale=0.5]}] (n11) -- coordinate[midway, yshift=\curveshift pt]({e112\i}) (n2\i) {};
        \pgfmathrandomitem{\mean}{means}\mean
        \pgfmathrandomitem{\std}{stds}\std
        \node[anchor=center] at (e112\i) {
            \adjustbox{max width=\curvesize pt}{\distplot{\mean}{\std}{#4}}
        };
    }
    
    \foreach \i in {1, ..., 3}{
        \draw[-{Stealth[scale=0.5]}] (n2\i) -- coordinate[midway, yshift=\curveshift pt]({e2\i3\i}) (n3\i) {};
        \pgfmathrandomitem{\mean}{means}\mean
        \pgfmathrandomitem{\std}{stds}\std
        \node[anchor=center] at (e2\i3\i) {
            \adjustbox{max width=\curvesize pt}{\distplot{\mean}{\std}{#4}}
        };
    }

    \foreach \i in {1, ..., 3}{
        \draw[-{Stealth[scale=0.5]}] (n3\i) -- coordinate[midway, yshift=\curveshift pt]({e3\i41}) (n41) {};
        \pgfmathrandomitem{\mean}{means}\mean
        \pgfmathrandomitem{\std}{stds}\std
        \node[anchor=center] at (e3\i41) {
            \adjustbox{max width=\curvesize pt}{\distplot{\mean}{\std}{#4}}
        };
    }
}

\begin{tikzpicture}[
    font=\footnotesize\sffamily,
    box/.append style={
        draw=none,
        fill=hgfgray!15!white,
        minimum width=0.4\linewidth, 
        minimum height=0.92\linewidth
    },
    topbox/.append style={
        draw=none, 
        minimum width=0.2\linewidth, 
        minimum height=0.25\linewidth,
        inner sep=0
    },
    label/.append style={
        draw=none,
        font=\normalsize\bfseries\sffamily,
        inner sep=0.8em
    },
]
    %% PROCESSOR 1
    \node[box] (proc1) {};
    \node[label, color=hgfblue, anchor=north west] (proc1-label) at (proc1.north west) {Processor 1};

    \node[topbox, below=2.5em of proc1.north west, anchor=north west] (proc1-left-box) {};
    \node[topbox, below=2.5em of proc1.north east, anchor=north east] (proc1-right-box) {};

    \node at (proc1-left-box) {\fontsize{26}{26}\selectfont\faDatabase};
    \node[anchor=south, above=0.0 of proc1-left-box.south] {\bfseries Data};

    \model{proc1-right-box.north}{0.2}{hgfmatter}
    \node[anchor=south, above=0.0 of proc1-right-box.south] {\bfseries Model};

    \node[topbox, below=1.0 of proc1-left-box] (proc1-model-box) {};
    \modelwithdists{proc1-model-box.north}{0.2}{hgfblue}{hgfmatter}
    \node[anchor=south, above=0.0 of proc1-model-box.south] {\bfseries Weight sample};

    \draw[-latex, shorten >=0.1cm] (proc1-left-box.south) -- (proc1-model-box.north) node[midway, right, align=left] {Sample\\ weights};

    \node[below=1.0 of proc1-model-box, color=hgfblue] (proc1-grad) {\Large $\nabla f_n$};
    \draw[-latex] (proc1-model-box.south) -- (proc1-grad.north) node[midway, right, align=left] {Forward and\\backward pass};
    
    \node[below=1.5 of proc1-right-box] (proc1-avg-grad) {\Large $\nabla f_{avg}$};
    \draw[-latex] (proc1-avg-grad.north) -- (proc1-right-box.south) node[midway, right, align=left] {Model\\update};

    %% PROCESSOR 2
    \node[box, right=1.8 of proc1] (proc2) {};
    \node[label, color=hgfinformation, anchor=north west] (proc2-label)  at (proc2.north west) {Processor n};

    \node[topbox, below=2.5em of proc2.north west, anchor=north west] (proc2-left-box) {};
    \node[topbox, below=2.5em of proc2.north east, anchor=north east] (proc2-right-box) {};

    \model{proc2-left-box.north}{0.2}
    \node[anchor=south, above=0.0 of proc2-left-box.south] {\bfseries Model};

    \node at (proc2-right-box) {\fontsize{26}{26}\selectfont\faDatabase};
    \node[anchor=south, above=0.0 of proc2-right-box.south] {\bfseries Data};

    \node[topbox, below=1.0 of proc2-right-box] (proc2-model-box) {};
    \modelwithdists{proc2-model-box.north}{0.2}{hgfinformation}{hgfmatter}
    \node[anchor=south, above=0.0 of proc2-model-box.south] {\bfseries Weight sample};

    \draw[-latex, shorten >=0.1cm] (proc2-right-box.south) -- (proc2-model-box.north) node[midway, left, align=left] {Sample\\ weights};

    \node[below=1.0 of proc2-model-box, color=hgfinformation] (proc2-grad) {\Large $\nabla f_n$};
    \draw[-latex] (proc2-model-box.south) -- (proc2-grad.north) node[midway, left, align=left] {Forward and\\backward pass};
    
    \node[below=1.5 of proc2-left-box] (proc2-avg-grad) {\Large $\nabla f_{avg}$};
    \draw[-latex] (proc2-avg-grad.north) -- (proc2-left-box.south) node[midway, left, align=left] {Model\\update};

    %% ARROWS
    \draw[draw=none] (proc1-grad) -- (proc2-grad) node[midway, fill=white, align=center] (allreduce) {\bfseries ALL\\\bfseries REDUCE};
    \draw[-latex, densely dotted, thick] (proc1-grad) -- (allreduce);
    \draw[-latex, densely dotted, thick] (proc2-grad) -- (allreduce);

    \draw[-latex, densely dotted, thick] ([xshift=-0.2cm, yshift=0.2cm]allreduce.north) |- (proc1-avg-grad.east);
    \draw[-latex, densely dotted, thick] ([xshift=0.2cm, yshift=0.2cm]allreduce.north) |- (proc2-avg-grad.west);
\end{tikzpicture}
    \caption{
        %Visualization of 
        BNN Training with Parallel Sampling \label{BNN_train}
    }
    \Description{}
\end{figure}
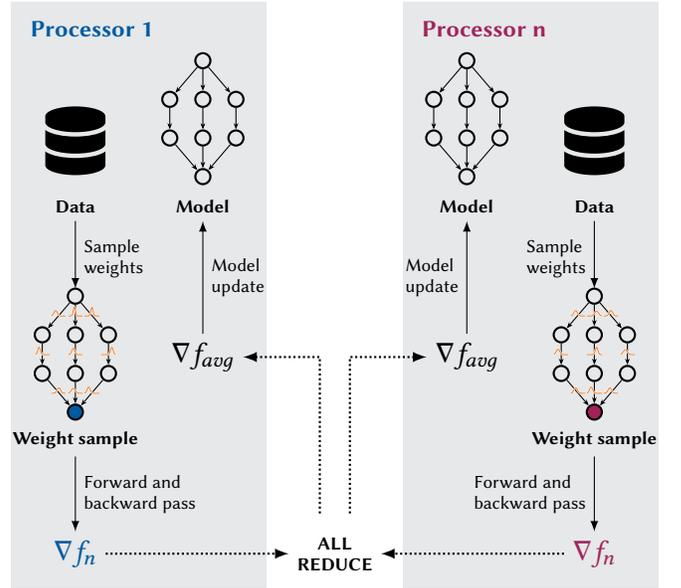

Our approach targets the reduction of the additional per-GPU memory demands and the parallelization of the computational burden by distributing precisely what makes sampling-based UQ methods more complex and memory demanding: the samples. \Cref{BNN_train} illustrates the underlying idea.

In our approach, the $s$ random parameter samples are distributed to the $p$ GPUs that are available for training. Thus, each GPU is responsible for drawing $s/p$ samples and performing the forward pass on them. This reduces the memory requirements of model training by a factor of up to $p$ as well as speeding up the training.

Algorithm \ref{BNN-algo} illustrates the implementation of sampling parallelism on the example of BNNs trained with VI, and highlights the differences to conventional BNN VI training (c.f. \ref{bnn-train-regular}).
Each GPU loads the same mini-batch from the dataset to ensure consistent data input across computations. 
While this replication introduces a significant overhead that initially limits scalability, it provides a distinct advantage: each duplicated batch can undergo random data augmentations, thereby enhancing augmentation diversity.

Sampling parallelism enables multiple stochastic forward passes to be performed on an identical model–dataset pair while employing distinct random seeds. These seeds govern not only the sampling of network parameters, but also the stochastic components of the data pipeline, including random data augmentations. Extensive prior work has demonstrated that increased diversity and frequency of data augmentation improve generalization performance ~\cite{brakel2024model}. Consequently, sampling parallelism effectively yields additional augmented training instances without incurring extra sequential training cost. Formally, each parallel run corresponds to a joint Monte Carlo draw from the distribution over model parameters and stochastic transformations.

Because BNNs represent model weights as probability distributions rather than point estimates, every forward pass requires drawing samples from these weight distributions. To properly capture their variability, multiple weight samples are drawn at every iteration during both training and inference. Accordingly, each GPU samples model weights using a unique random seed, ensuring that no redundant samples are generated.
Each GPU then performs a forward pass to compute its local predictions. The ELBO loss for the current batch is evaluated independently on each GPU, and the corresponding gradients are computed locally. Finally, an allreduce operation is performed to average and synchronize the gradients across GPUs, after which the model parameters are updated. This synchronization ensures that all GPUs remain consistent and that the updates reflect contributions from multiple independent weight samples, thereby reducing gradient variance and stabilizing training.

\begin{algorithm}%[H]
\caption{Sampling Parallel Bayesian Neural Network Training with Variational Inference \label{BNN-algo}}
\begin{algorithmic}[1]
\Require Dataset $\mathcal{D}$, number of epochs $E$, number of samples $S$
\State Initialize variational parameters $\mu$ and $\sigma$ for each weight % $\sigma = \text{softplus}(\rho)$
\For{$\text{epoch} = 1$ to $E$}
    \For{each minibatch $(x, y) \in \mathcal{D}$}
    \For {\hl{each GPU $p$ in parallel}}
        \For {$\text{s} = 1$ to $S/P$}
        \State Sample $\epsilon_s \sim \mathcal{N}(0, I)$
        \State $w_{p,s} \gets \mu + \sigma \odot \epsilon_s$ \hfill %// reparameterization trick
        \State $\hat{y}_{p,s} \gets \text{ForwardPass}(x, w_{p,s})$
        \State $\mathcal{L}_{\text{p,s,data}} \gets \text{Loss}(\hat{y}_{p,s}, y)$ \hfill %// NLL / CE / MSE
        \State $\mathcal{L}_{\text{KL}} \gets \frac{1}{2} \sum_i \big(\sigma_i^2 + \mu_i^2 - 1 - \log \sigma_i^2 \big)$% // prior matching
        \State $\mathcal{L}_{p,s} \gets \mathcal{L}_{\text{p,s,data}} + \frac{1}{|\mathcal{D}|} \mathcal{L}_{\text{KL}}$ \hfill %// negative scaled ELBO        
        \State Compute gradients $\nabla \mathcal{L}_s(\mu, \sigma)$
        \EndFor
        \State \hl{All-reduce communication from all GPUs}
        \State Average gradients $\frac{1}{S}\sum_{p=1}^{P}\sum_{s=1}^{S/p}\nabla \mathcal{L}_{p,s}(\mu, \sigma)$
        \State Update $\mu$ and $\sigma$ using optimizer (e.g., Adam)
        \EndFor
    \EndFor
\EndFor
\end{algorithmic}
\end{algorithm}

It is important to note that the distributed implementation is not an exact replica of the non-distributed algorithm.
While the prior matching part of the ELBO loss can be calculated and aggregated without additional approximations by simply averaging the gradients between GPUs, the data term in general can depend on non-linear functions of, e.g., mean and standard deviation of the samples. However, when these samples are distributed across multiple GPUs, obtaining these statistics becomes nontrivial. In particular, standard deviations cannot be correctly aggregated through simple averaging; thus, relying solely on a gradient averaging yields only an approximation. 

A simple example would be the case of model trained on a classification task with a cross-entropy loss on the (arithmetic) mean of the class probabilities of the individual predictions.
In the extreme case of the $s$ samples being distributed onto $p=s$ GPUs, each with a single prediction, the effective loss would correspond to aggregating the class probabilities with a geometric mean instead.

Despite this discrepancy, the distributed implementation with only gradient averaging remains an approximation of the local algorithm, and our empirical results indicate that it exhibits very similar behavior in practice. 
%An exact implementation would be possible by communicating the predictions themselves rather than the gradients. 
This approach allows us to rely on existing, highly optimized frameworks that handle gradient communication and synchronization automatically. %We return to this point in the discussion of our results.

Nonetheless, an exact algorithm where the standard deviation and mean are aggregated across multiple GPUs is also possible. This requires the additional communication of two parameters.

\subsection{Combining Sampling Parallelism with DDP \label{sec:hybrid}}

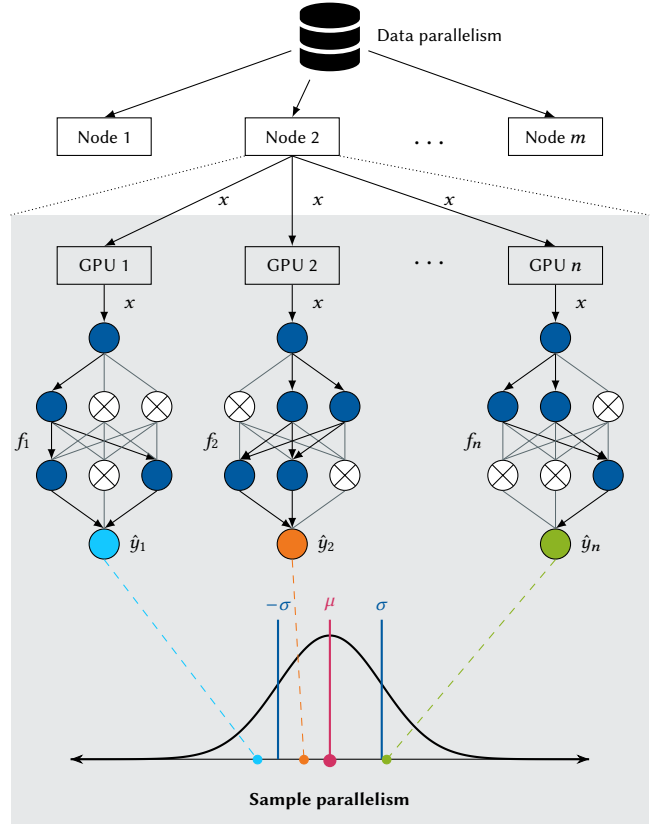
\begin{figure}
    \centering
    \pgfmathdeclarefunction{gauss}{2}{%
    \pgfmathparse{1/(#2*sqrt(2*pi))*exp(-((x-#1)^2)/(2*#2^2))}%
}

\begin{tikzpicture}[
    font=\footnotesize\sffamily,
    server/.append style={
        draw=black,
        minimum width=1.25cm,
        minimum height=0.5cm,
        anchor=north
    },
    neuron/.append style={
        fill=hgfblue,
        circle,
        draw=black,
        inner sep=0pt, 
        minimum width=0.4cm,
        anchor=north
    },
    dead-neuron/.append style={
        fill=white,
        circle,
        draw=black,
        inner sep=0pt, 
        minimum width=0.4cm,
        path picture={
            \draw[black] 
                (path picture bounding box.north west) -- (path picture bounding box.south east)
                (path picture bounding box.north east) -- (path picture bounding box.south west);
        },
        anchor=north
    },
    active/.append style={
        -latex
    },
    dead/.append style={
        -, 
        hgfgray, 
        ultra thin
    },
    gpu/.append style={
        thick, 
        draw=black,
        minimum width=0.7cm,
        minimum height=0.7cm,
    }
]
    \node (data) {\fontsize{26}{26}\selectfont\faDatabase};
    \node[right=0.0 of data] {Data parallelism};

    \node[below left=0.5 and 3.0 of data.south, server] (n1) {Node 1};
    \node[below left=0.5 and 0.5 of data.south, server] (n2) {Node 2};
    \node[below right=0.5 and 3.0 of data.south, server] (n4) {Node $m$};

    \draw[-latex] (data) -- (n1.north);
    \draw[-latex] (data) -- (n2.north);
    \draw[-latex] (data) -- (n4.north);

    \node[draw=none, fill=hgfgray!15!white, minimum width=\linewidth, minimum height=8.1cm, anchor=north, below=1.8 of data] (node-box) {};
    \draw[densely dotted] (n2.south west) -- (node-box.north west);
    \draw[densely dotted] (n2.south east) -- (node-box.north east);
    
    \node[anchor=south, above=0.1 of node-box.south] {\bfseries Sample parallelism};
    
    \node[below=1.2 of n1.south, server] (g1) {GPU 1};
    \node[below=1.2 of n2.south, server] (g2) {GPU 2};
    \node[below=1.2 of n4.south, server] (g4) {GPU $n$};

    % left nn
    \node[neuron, fill=hgfblue, below=0.5 of g1.south] (n111) {};
    
    \node[below left=0.5 and 0.7 of n111.south, neuron] (n121) {};
    \node[below=0.5 of n111.south, dead-neuron] (n122) {};
    \node[below right=0.5 and 0.7 of n111.south, dead-neuron] (n123) {};

    \draw[active] (n111.south) -- (n121.north);
    \draw[dead] (n111.south) -- (n122.north);
    \draw[dead] (n111.south) -- (n123.north);
    
    \node[neuron, below=0.5 of n121.south] (n131) {};
    \node[dead-neuron, below=0.5 of n122.south] (n132) {};
    \node[neuron, below=0.5 of n123.south] (n133) {};

    \draw[active] (n121.south) -- (n131.north) node[midway, left=0.5em, black] {$f_1$};
    \draw[dead] (n121.south) -- (n132.north);
    \draw[active] (n121.south) -- (n133.north);
    
    \draw[dead] (n122.south) -- (n131.north);
    \draw[dead] (n122.south) -- (n132.north);
    \draw[dead] (n122.south) -- (n133.north);
    
    \draw[dead] (n123.south) -- (n131.north);
    \draw[dead] (n123.south) -- (n132.north);
    \draw[dead] (n123.south) -- (n133.north);
    
    \node[neuron, fill=hgflightblue, below=0.5 of n132.south] (n141) {};
    \node[right=0.0 of n141] {$\hat{y}_1$};
    
    \draw[active] (n131.south) -- (n141.north);
    \draw[dead] (n132.south) -- (n141.north);
    \draw[active] (n133.south) -- (n141.north);

    % second left nn
    \node[neuron, below=0.5 of g2.south] (n211) {};
    
    \node[below left=0.5 and 0.7 of n211.south, dead-neuron] (n221) {};
    \node[below=0.5 of n211.south, neuron] (n222) {};
    \node[below right=0.5 and 0.7 of n211.south, neuron] (n223) {};

    \draw[dead] (n211.south) -- (n221.north);
    \draw[active] (n211.south) -- (n222.north);
    \draw[active] (n211.south) -- (n223.north);
    
    \node[neuron, below=0.5 of n221.south] (n231) {};
    \node[neuron, below=0.5 of n222.south] (n232) {};
    \node[dead-neuron, below=0.5 of n223.south] (n233) {};

    \draw[dead] (n221.south) -- (n231.north) node[midway, left=0.5em, black] {$f_2$};
    \draw[dead] (n221.south) -- (n232.north);
    \draw[dead] (n221.south) -- (n233.north);
    
    \draw[active] (n222.south) -- (n231.north);
    \draw[active] (n222.south) -- (n232.north);
    \draw[dead] (n222.south) -- (n233.north);
    
    \draw[active] (n223.south) -- (n231.north);
    \draw[active] (n223.south) -- (n232.north);
    \draw[dead] (n223.south) -- (n233.north);
    
    \node[neuron, fill=hgfmatter, below=0.5 of n232.south] (n241) {};
    \node[right=0.0 of n241] {$\hat{y}_2$};
    
    \draw[active] (n231.south) -- (n241.north);
    \draw[active] (n232.south) -- (n241.north);
    \draw[dead] (n233.south) -- (n241.north);

    % dots
    \node[below right=0.7 and 1.0 of data.south] (n3) {\Large $\dots$};
    \node[below right=2.3 and 1.0 of data.south] (n311) {\Large $\dots$};

    % right nn
    \node[neuron, below=0.5 of g4.south] (n411) {};
    
    \node[fill=hgfblue, below left=0.5 and 0.7 of n411.south, neuron] (n421) {};
    \node[below=0.5 of n411.south, neuron] (n422) {};
    \node[below right=0.5 and 0.7 of n411.south, dead-neuron] (n423) {};

    \draw[active] (n411.south) -- (n421.north);
    \draw[active] (n411.south) -- (n422.north);
    \draw[dead] (n411.south) -- (n423.north);
    
    \node[dead-neuron, below=0.5 of n421.south] (n431) {};
    \node[dead-neuron, below=0.5 of n422.south] (n432) {};
    \node[neuron, below=0.5 of n423.south] (n433) {};

    \draw[dead] (n421.south) -- (n431.north) node[midway, left=0.5em, black] {$f_n$};
    \draw[dead] (n421.south) -- (n432.north);
    \draw[active] (n421.south) -- (n433.north);
    
    \draw[dead] (n422.south) -- (n431.north);
    \draw[dead] (n422.south) -- (n432.north);
    \draw[active] (n422.south) -- (n433.north);
    
    \draw[dead] (n423.south) -- (n431.north);
    \draw[dead] (n423.south) -- (n432.north);
    \draw[dead] (n423.south) -- (n433.north);
    
    \node[neuron, fill=hgfgreen, below=0.5 of n432.south] (n441) {};
    \node[right=0.0 of n441] {$\hat{y}_n$};
    
    \draw[dead] (n431.south) -- (n441.north);
    \draw[dead] (n432.south) -- (n441.north);
    \draw[active] (n433.south) -- (n441.north);
    
    \begin{axis}[
        at={(0.0, -7.8cm)},
        anchor=north,
        width=1.0\linewidth,
        height=0.4\linewidth,
        axis x line=bottom,
        x axis line style={
            draw=black,
            Stealth-Stealth
        },
        axis y line=none,
        xlabel=,
        ylabel=,
        ymin=0.0,
        clip=false
    ]
        \addplot[black, mark=none, smooth, samples=100, thick] {gauss(0, 1.0)};
        
        \draw[hgfblue, thick] (axis cs:-1.0, 0.0) -- (axis cs:-1.0, 0.45) node[above] {$-\sigma$};
        \draw[Circle-, hgfhealth, thick, shorten <=-3pt] (axis cs:0.0, 0.0) -- (axis cs:0.0, 0.45) node[above] {$\mu$};
        \draw[hgfblue, thick] (axis cs:1.0, 0.0) -- (axis cs:1.0, 0.45) node[above] {$\sigma$};
        
        \coordinate (point-a) at (axis cs: -1.4, 0.0);
        \coordinate (point-b) at (axis cs: -0.5, 0.0);
        \coordinate (point-c) at (axis cs: 1.1, 0.0);
    \end{axis}

    \draw[-Circle, dashed, hgflightblue, shorten >=-2pt] (n141.south) -- (point-a);
    \draw[-Circle, dashed, hgfmatter, shorten >=-2pt] (n241.south) -- (point-b);
    \draw[-Circle, dashed, hgfgreen, shorten >=-2pt] (n441.south) -- (point-c);
    
    \draw[-latex] (n2.south) -- (g1.north) node[midway, right=0.5em] {$x$};
    \draw[-latex] (n2.south) -- (g2.north) node[midway, right=0.5em] {$x$};
    \draw[-latex] (n2.south) -- (g4.north) node[midway, right=0.5em] {$x$};

    \draw[-latex] (g1) -- (n111) node[midway, right=0.5em] {$x$};
    \draw[-latex] (g2) -- (n211) node[midway, right=0.5em] {$x$};
    \draw[-latex] (g4) -- (n411) node[midway, right=0.5em] {$x$};
\end{tikzpicture}
    \caption{Distributing Training with Hybrid Parallelization \label{hybrid}}
    \Description{}
\end{figure}

%The goal of our approach is not to replace or improve on existing parallelization techniques, but to 
Sampling parallelism aims to tackle the root cause of computation and memory demand specifically introduced by the sampling-based nature of Bayesian learning techniques. As such, it adds an additional axis for parallelization that can be leveraged orthogonally to existing approaches. We illustrate this, by combining our baseline sampling parallelism algorithm with Distributed Data Parallelism (DDP) in a hybrid parallelization strategy. In this setup, weight samples are distributed across GPUs within the same node, while the data are distributed across multiple nodes. This design allows us to leverage the strengths of both approaches.
\Cref{hybrid} shows how parallelization is handled in this hybrid parallelization.

\section{Experimental Setup}
We evaluate the proposed algorithms and demonstrate their benefits, feasibility, and scalability using three different use cases. For clarity, we focus primarily on BNNs trained with VI, though the underlying concepts and methods readily extend to other sampling-based Bayesian learning approaches, datasets, architectures, and tasks. We demonstrate this by extending the experiments to a use case using MCD.

\subsection{Hardware and Software}
All experiments were performed on a high-performance computing cluster, where each node features 4 NVIDIA 40GB A100 GPUs connected via NVLink3, and 2 AMD EPYC 7402 CPUs with 512 GB RAM, managed through Slurm.

All experiments were conducted using \texttt{Python 3.12.3} and \texttt{CUDA 12.8}. We employed \texttt{PyTorch 2.9.0}~\cite{paszke2019pytorch} and \texttt{torch\_blue}~\cite{rai_scc_torch_blue} as the primary frameworks for Bayesian learning, and used \texttt{torch.\allowbreak distributed} together with \texttt{Distributed\allowbreak Data\allowbreak Parallel}~\cite{li2020pytorch} for parallelization.

\subsection{Use Case 1: ViT on CIFAR-10}
The primary use case we evaluate our approach on is the task of image classification.
We employ a Bayesian version of a Vision Transformer (ViT) architecture~\cite{dosovitskiy2020image}, and train it with sampling-parallel variational inference.
The ViT architecture features $4\times4$ patches and an embedding dimension of 192 with three attention heads (64 dimensions per head). The Transformer encoder contains six layers, each followed by an MLP block with a hidden size of 768 (a 4$\times$ expansion). To model parameter uncertainty, all weights are assigned a mean-field normal variational distribution with a corresponding mean-field normal prior. Parameter means are initialized using Kaiming initialization and variances are initialized using a constant initialization scaled inversely with the layer width. The variational parameters are optimized via standard VI with Bayes-by-Backprop~\cite{blundell2015weight}, maximizing the ELBO with a categorical predictive distribution for classification. The ViT model outputs one set of class probabilities for each sampling of its weights.

As dataset, we use \textbf{CIFAR-10} consisting of 60,000 images ($32\times32$ pixels each), with the canonical 50k/10k train/test split, while reserving 10\% of the training set for validation. 

We apply a standard set of data augmentations:
Each image is randomly cropped to $32\times32$ with a padding of 4 pixels, then horizontally flipped with a probability of 50\%. 
Images are converted to tensors and normalized using the CIFAR10 mean and standard deviation for each channel.
During validation and testing, the predictive performance of the ViT is assessed using top-1 classification accuracy computed from the model’s predictive mean.

\subsection{Use Case 2: Time-Series Forecasting with an MLP}
To illustrate the generalizability of our approach towards all sampling-based Bayesian learning methods, we further explore its application to MCD. For this use case we chose the task of time-series forecasting, using a simple Multilayer Perceptron (MLP) with two hidden layers, each featuring a width of 128 neurons. The choice of a comparatively small model allows us to further study the speedup and efficiency of sampling parallelism in scenarios in which the GPUs are not fully utilized. 

We train the Bayesian MLP using the \textbf{ENTSO-E de} dataset~\cite{ENTSOE_Germany_Load}, which contains electricity consumption data for Germany in 15 min intervals over 5 years. The forecasting task is to predict the electricity consumption of the next 6 hours (24 data points) given the consumption of the previous 24 hours (96 data points). The MLP outputs a set of scalar predictions for each random sample and each of the 24 target time points and is trained on the Mean Squared Error (MSE) loss of the averaged predictions.
We also use the MSE of the average predictions for the purposes of validation and testing.

%\subsection{Architecture \label{sec:architecture}}
\subsection{Use-Case 3: Bayesian Weather Models}
While use cases one and two allow us to distinctly study performance and scaling behaviour of our sampling parallelism approach, they are ultimately too small to fully highlight its unique advantages.
In particular, for the Bayesian ViT, the increased computational load and memory demand can be tackled using simple DDP: By decreasing the local batch size per GPU, more memory becomes available for the model and its samples without decreasing efficiency and GPU utilization. While this approach is technically limited by the minimum local batch size of one sample, the comparatively small size of the CIFAR-10 data items makes this limit practically impossible to reach. 

However, this is not the case for applications where individual data points themselves are very large. A paradigm in this regard is data-driven weather forecasting, where data items consist of high-resolution global maps of atmospheric state variables. The field is currently heavily researched and the immense sizes of data and models alike are pushing the boundaries of what is capable with modern accelerator hardware. 
Already now, state-of-the-art networks can often only be trained using both model and data parallel approaches, with the latter operating under the constraint that only one or two data items fit into accelerator memory.
Moreover, these models tend to exhibit large batch effects rather quickly (around an effective global batch size of 8 to 16, according to our experience), which inherently limits the scalability of DDP. This makes the application of sampling-based Bayesian learning methods in these models, in particular BNN versions thereof, virtually impossible.
\begin{figure*}
    \centering
    \input{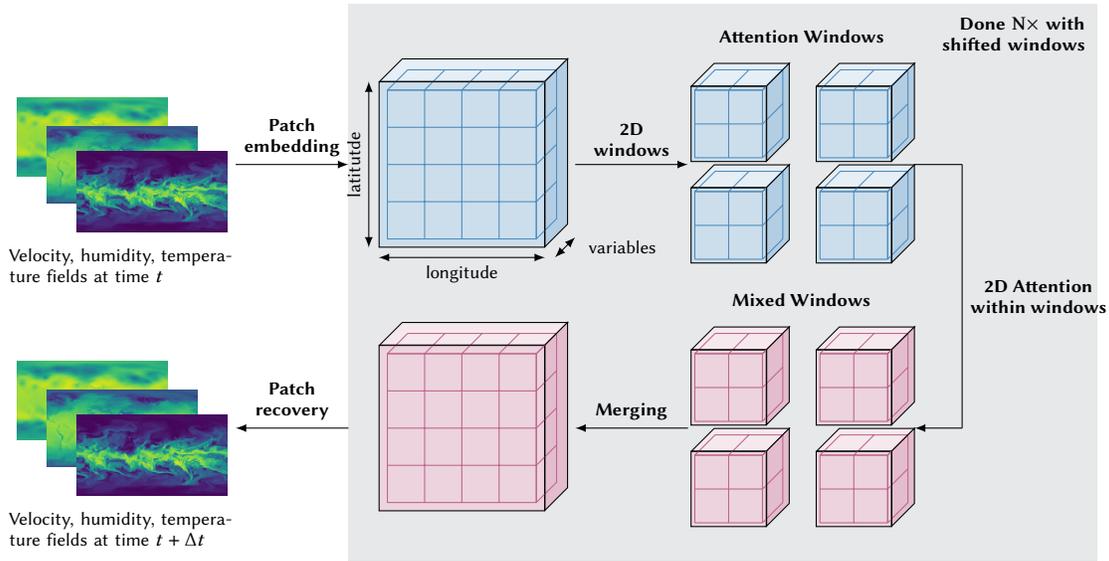}
    \caption{Simplified illustration of the SWIN transformer architecture used for the weather forecasting task; patch embedding and recovery is performed via a 2d convolution or transpose convolution layer with kernel size = stride = $(2, 2)$ over the latitude and longitude dimensions}
    \label{fig:kangu-weather}
    \Description{}
\end{figure*}

To illustrate how our approach of sampling parallelism can aid in overcoming this challenge, we examine this application in a third use case. 
The intention of this evaluation is to function primarily as a representative for large regression models that are typically too big to make Bayesian, rather than as a model tailored for this specific task. Using sampling parallelism, we demonstrate that sampling-based Bayesian learning can be applied to models of this scale.

We use a simple Shifted Window (SWIN) Transformer architecture based on~\citet{to2024architectural} without the learned positional embedding for the task of weather forecasting. The patch size is set to $2\times2$ and the embedding dimension to $540$.
We use 6 SWIN blocks with 12 heads each before and after down- and upsampling and 4 SWIN blocks with 24 heads each in-between down- and upsampling. \Cref{fig:kangu-weather} shows the architecture of this model.
It has over 200 million parameters, and each data item is roughly 17.7 MBs.

To model parameter uncertainty, all weights of the model are assigned a mean-field normal variational distribution with a corresponding mean-field normal prior. The network is initialized using Kaiming initialization for the means and a constant initialization scaled inversely with the layer width for the variances. The variational parameters are optimized via standard VI with Bayes-by-Backprop~\cite{blundell2015weight}. 
%However, this model is intended more as a readily available stand-in for large regression models in the field in terms of model size than as a SOTA model for weather prediction in terms of predictive quality.
We train the model on the \textbf{ERA5} dataset~\cite{hersbach2020era5} at $1.5^\circ$ resolution (downloaded via Weatherbench 2 cloud storage~\cite{rasp2024weatherbench}), restricted to three-hour subsampling (i.e., only 00:00 UTC, 03:00 UTC, etc.) and data from 1980 to 2020. The task is a 6h forecast of the same 69 variables used in the Pangu-Weather model~\cite{bi2023accurate}. We also add the same constant masks as part of the input as well as longitude and latitude features combined with time-of-day and day-of-year information respectively, resulting in data samples with dimensions $76\times121\times240$. The data is augmented by random periodic shifts in longitude.
The weather SWIN Transformer outputs a set of forecasts for each variable and grid point. 
We train the model by maximizing the ELBO, using a Gaussian negative log-likelihood for the data term in ELBO. 
Validation and testing are performed via latitude-weighted RMSEs of individual variables. However, since predictive performance was not the focus of this, these models were not trained to convergence and thus the RMSEs are not reported.

All source code and evaluation scripts used in our experiments is available open-source\footnote{Will be made publicly available upon publication}.

\section{Results \& Discussions}
To evaluate the feasibility and efficiency of our proposed parallelization scheme, we perform a series of scaling experiments in which we observed both the runtime as well as the convergence of accuracy values. 
We conduct two types of experiments. For one, we explore scaling experiments in which the workload is scaled proportionally to the increasing resources to establish that our parallelization scheme works reasonably well. This type of scaling experiment, which we will be calling proportional-sample scaling, evaluates how runtime changes when both the problem size and the number of GPUs grow proportionally, keeping the workload per GPU constant. In our case the workload is determined by the number of samples drawn. \emph{Efficiency} is then evaluated as 
$E(p) = \frac{T_{1}(n)}{T_{p}(n \cdot p)}$.
In our proportional-sample scaling experiments, we start with a non-Bayesian model on a single GPU and increase the number of samples drawn proportional to the number of GPUs, keeping every other hyperparameter fixed. 

Proportional-sample scaling highlights how an algorithm behaves when parallel workload grows to exclude the diminishing returns associated with fixed-size problems. However, ultimately, the goal of parallelization is to accelerate computation of a fixed workload by adding additional compute resources, i.e., fixed-sample scaling.
Fixed-sample scaling measures how the runtime of a fixed-size problem improves as the number of processing units increases. 
Its primary evaluation metric is \emph{speed-up}, defined as the ratio of the single-GPU runtime to the runtime using $p$ GPUs $
S(p) =\frac{T_{1}(n)}{T_{p}(n)}$,
where $T_1(n)$ is the required time for a task of size $n$ on a single GPU and $T_p(n)$ is the corresponding cost when the same task is distributed onto $p$ GPUs.
This metric characterizes how effectively additional compute resources reduce computation time and reveal upper limits imposed by the serial portion of the workload.

For our fixed-sample scaling experiments, we increase the number of GPUs while keeping the workload (number of random samples) constant. There are two variations on how we conduct these experiments. In one setup, we do not change any other hyperparameter. While this enables the model to maintain the same setup, it effectively reduces the load per GPU making memory and time usage inefficient. In the other case, we ensure GPU usage at maximum capacity by increasing the batch size proportionally. 
For comparative reasons, we also parallelize the same model via DDP. Again we study two cases, to match our experiments with sampling parallelism. For DDP, the equivalent for case 1 is, to keep global batch size constant, meaning the batch size per GPU decreases. For the second case, we keep the local batch size constant, ensuring that GPUs run at capacity. As for sampling and hybrid parallelism, we also investigate the two variants that maintain either the global batch size or the GPU load. 

\subsection{ViT on CIFAR-10}
We start by evaluating the efficiency of sampling parallelism by conducting proportional-sample scaling on the task of image classification, using a Bayesian ViT on CIFAR-10. In a non-distributed setting, an epoch with 16 samples takes about 1,5 minutes for this task. 

\begin{figure}[ht]
    \centering
    \includegraphics[scale=0.4]{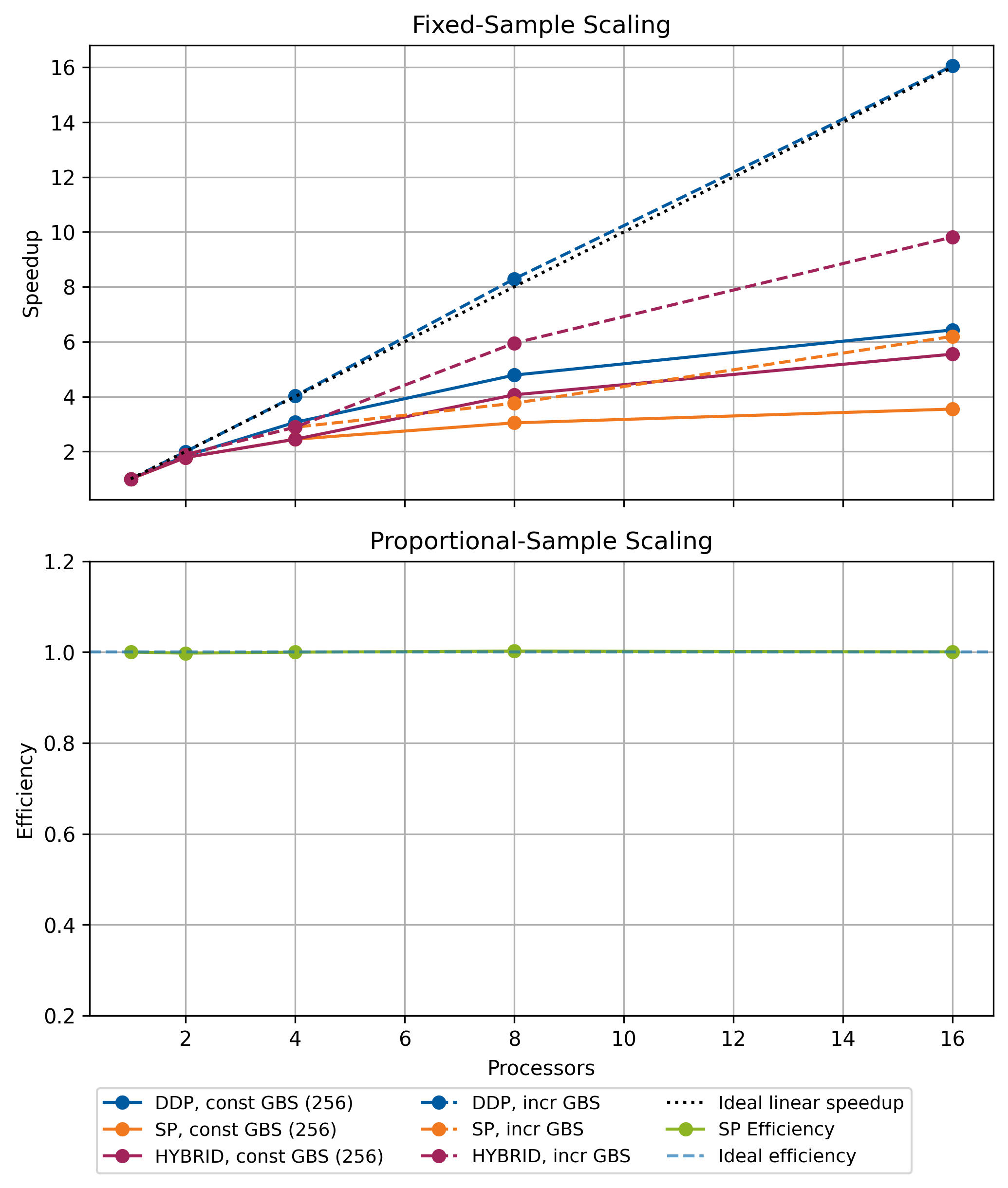}
    %\caption{Speed up (strong scaling, top) and efficiency (weak scaling, bottom) of training a Bayesian ViT on CIFAR10. For strong scaling, 16 random samples were drawn, and a global batch size of 256 (solid) and 256 $\times$ \# GPU (dashed) was used. For weak scaling \# Random Samples = \# GPU, with a global batch size of 1024.}
    \caption{Speed-up (fixed-sample scaling, top) and efficiency (proportional-sample scaling, bottom) of training the Bayesian ViT on CIFAR10. For fixed-sample scaling, we compare different sampling parallelism, data parallelism, and a hybrid of both, using 16 random samples and either a global batch size of 256 or an increasing global batch size of 256 times number of GPUs.
    For proportional-sample scaling we use a fixed global batch size of 1024 and scale the number of random samples from 1 to 16.}
    \label{fig:ViT-scaling}
    \Description{}
\end{figure}

%The global batch size was fixed to 1024.
The results depicted in~\Cref{fig:ViT-scaling} (bottom) show that the approach achieves near ideal scaling behavior; efficiency remains effectively constant over an increasing number of GPUs, with only minor point-to-point fluctuations attributable to measurement noise rather than algorithmic overhead. This near-perfect scaling indicates that the sampling procedure parallelizes cleanly, with no observable degradation in performance as the number of samples (and GPUs) increases. 
This is unsurprising since the task of distributed sampling in the way we propose it is embarrassingly parallel.
Consequently, the method is well-suited for large-scale sampling-based uncertainty estimation approaches in which many independent samples must be evaluated in parallel.

Another important point to emphasize is that increasing the number of drawn random samples from 1 to 16 would not have been feasible without the proposed parallelization scheme, due to memory limitations, unless the batch size was  altered as well. While alternative strategies, such as parallelizing other components of the model (tensor parallelism, pipelining etc.) to free up additional memory and compute time, could in principle enable larger sample counts, these approaches are considerably more intrusive and difficult to apply in practice. In contrast, our method provides a straightforward and scalable mechanism for distributing sample evaluations across multiple GPUs, enabling substantial increases in sampling throughput with minimal changes to the underlying model.

For the fixed-sample scaling experiments, we fixed the total number of samples to 16. As noted previously, this configuration cannot be executed on a single GPU without modification, so we reduced the local batch size to at most 256 to ensure that the experiment remained feasible under the same hardware constraints. 
All other settings were kept identical to the proportional-sample scaling setup, as well as between the different methods.

We then conduct an inter-parallelism comparison between data parallelism (DDP), sampling parallelism, and the hybrid approach combining both (see~\Cref{sec:hybrid}). The hybrid parallelization uses sampling parallelism for intra-node distribution and DDP for inter-node. For each strategy, we conducted the experiment in two variants: one in which the global batch size is held constant across GPU count, meaning that the local batch size and therefore load per GPU decreases, and one in which the global batch size is increased so that each device operates at full capacity. The achieved speed-ups based on the recorded per epoch run-times are summarized in~\Cref{fig:ViT-scaling}. When operating at capacity, DDP achieves near-ideal speedup, whereas hybrid and sample-parallel configurations fall short of perfect scaling. This gap arises because, in sampling parallelism, the same data batch must be loaded on every GPU, whereas DDP distributes different microbatches across devices. Since data loading constitutes a significant portion of the runtime, the duplicated data loading in sampling parallelism naturally limits the achievable speedup.

\begin{figure}[ht]
    \centering
    \includegraphics[scale=0.35]{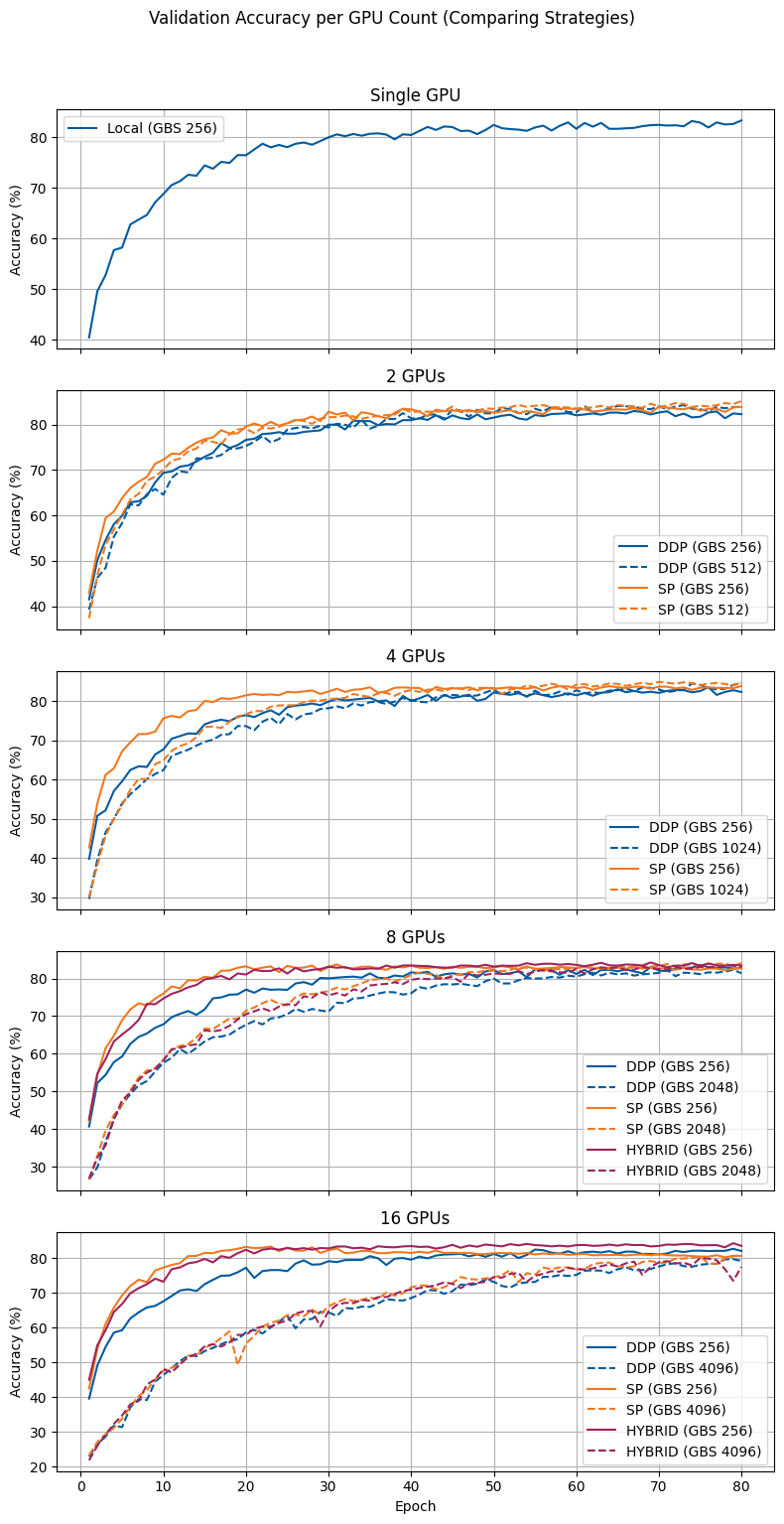}
    \caption{Accuracy of the Bayesian ViT on CIFAR10 with two different global batch sizes and 16 random samples, using sampling parallelism, DDP and hybrid parallelism, over the course of training,  for different numbers of GPUs.}
    \label{fig:ViT-strong-accuracy}
    \Description{}
\end{figure}

However, as outlined before, the duplicated data loading in sampling parallelism allows us to apply independent stochastic data augmentations locally per sample, thus increasing the augmentation diversity.
By loading the same data batch on all GPUs, we can apply different random augmentations, such as random crops and flips, on each of our GPUs for the same batch. 
When examining the corresponding convergence of the validation accuracy (c.f.~\Cref{fig:ViT-strong-accuracy}), we find that sampling parallelism converges substantially faster than DDP when per epoch accuracy increases are considered. 

\begin{figure}[ht]
    \centering
    \includegraphics[scale=0.3]{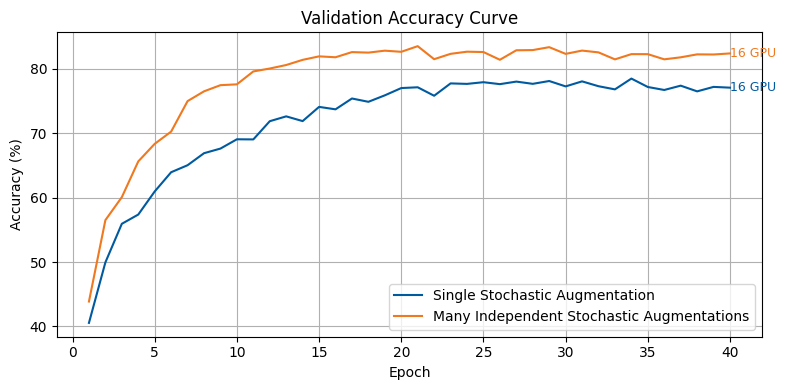}
    \caption{Accuracy of sampling parallelism over the course of training, using 16 random samples on 16 GPUS and two data augmentation settings: one in which all GPUs apply identical augmentations (blue), and one in which each GPU applies its own stochastic augmentations (orange). }
    \label{fig:ViT-augmentation-accuracy}
    \Description{}
\end{figure}

We investigate this effect further by analyzing the convergence behavior of sampling parallelism under two augmentation settings: one in which all GPUs apply identical augmentations, and one in which each GPU applies its own stochastic augmentations. The results depicted in~\Cref{fig:ViT-augmentation-accuracy} show that convergence is noticeably faster when GPUs perform independent augmentations.
These findings strengthen the usefulness of sampling parallelism. 
By increasing the variation of augmentations, sampling parallelism can effectively enlarge the dataset seen during training, enabling faster convergence and better generalisability.
While this effect could in principle also be replicated on the individual minibatch of a single GPU in DDP, this would require replicating the loaded data before augmentation, incurring additional compute and memory costs compared to sampling parallelism.

\begin{figure}[ht]
    \centering
    \includegraphics[scale=0.35]{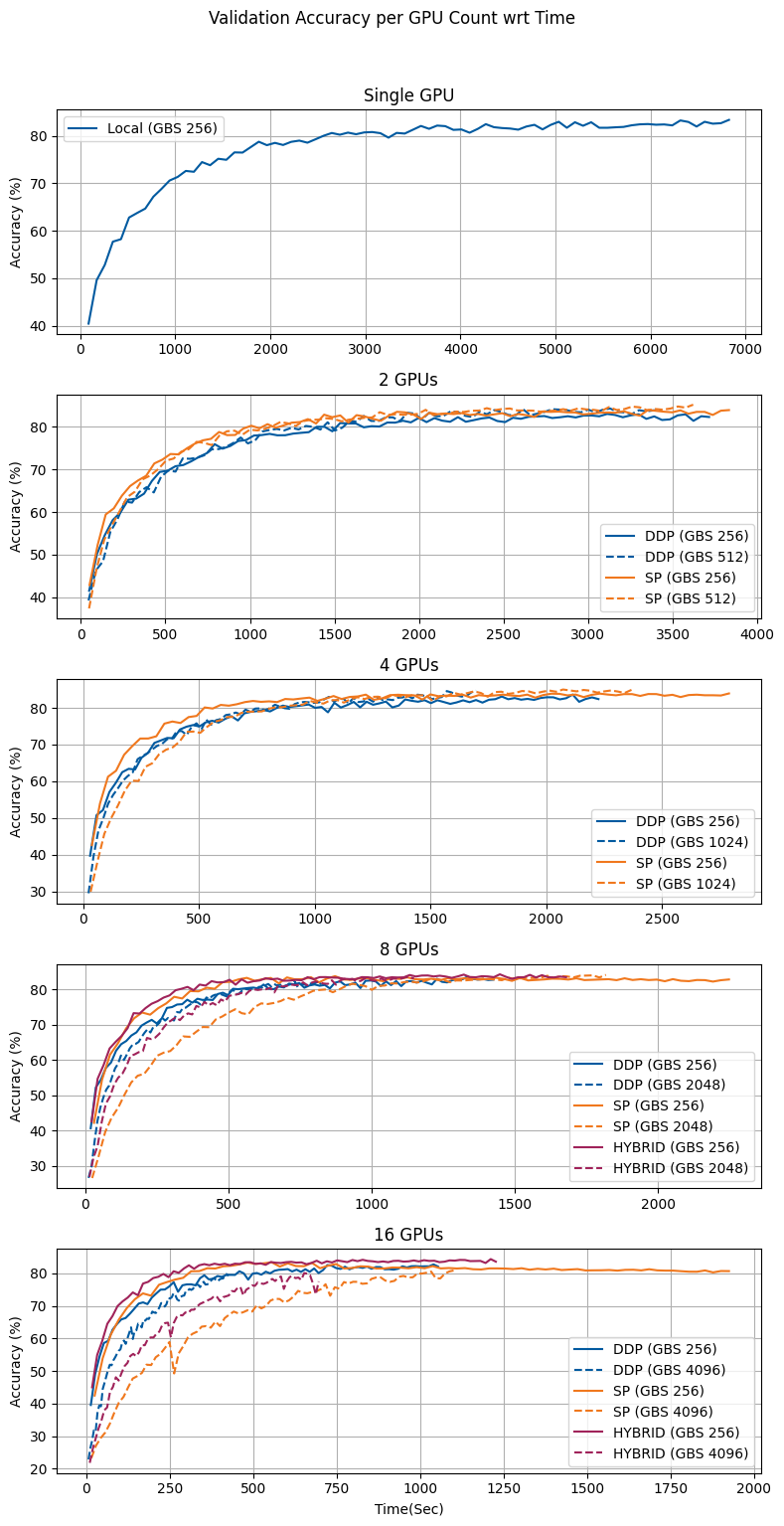}
    \caption{
    Accuracy of the Bayesian ViT on CIFAR10 with two different global batch sizes and 16 random samples, using sampling parallelism, DDP, and hybrid parallelism, with respect to wall-clock time,  for different numbers of GPUs.}
    \label{fig:time_accuracy}
    \Description{}
\end{figure}

Although DDP is more efficient in terms of epoch time, sampling parallelism requires fewer epochs to reach comparable accuracy. To assess overall convergence speed, we therefore compare accuracy as a function of wall-clock time, shown in~\Cref{fig:time_accuracy}. When viewed in this manner, the total time to reach a target accuracy is similar across all methods, with sampling parallelism showing a slight advantage due to its accelerated convergence in terms of epochs.

Taken together, these fixed-sample scaling results highlight an important trade-off between computational efficiency and convergence behavior. While DDP is the most effective approach for minimizing per-epoch runtime and achieves the best raw speedup under full device utilization, sampling parallelism's ability to leverage augmentation diversity across GPUs enables significantly faster learning per epoch, which compensates for its weaker scaling properties. Consequently, despite its higher data-loading overhead, sampling parallelism can match, and in some cases surpass, DDP in terms of wall-clock time to reach a target accuracy.

\begin{figure}[ht]
    \centering
    \includegraphics[scale=0.35]{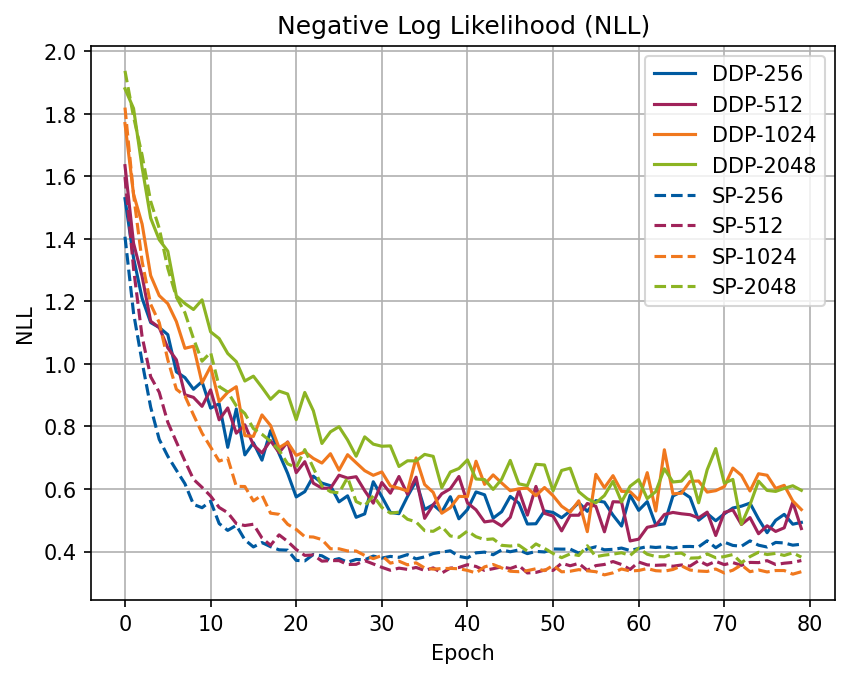}
    \caption{Negative log likelihood during training of the ViT on CIFAR-10 trained on 16 GPUs with 16 samples labeled with parallelization strategy and global batch size.}
    \label{fig:nll}
\end{figure}

To evaluate the impact not only on the model predictive performance, but also the quality of the uncertainty quantification, we examine the negative log likelihood (NLL) and the mean absolute calibration error as measures for uncertainty. \Cref{fig:nll} shows the progression of the NLL over the epochs of training with different global batch sizes as well as different parallelization strategies. While NLL decreases for all configurations (see~\Cref{fig:nll}), the sample parallel implementation yields better performance, due to the benefit of enhanced data augmentation. \Cref{tab:mace_results} shows the mean absolute calibration error, which is a measure of how well predicted uncertainty levels match the empirical errors of the model. Sampling parallelism yields consistently lower MACE, compared to DDP.

\begin{table}[ht]
\centering
\caption{Mean Absolute Calibration Error (MACE $\downarrow$) of the ViT on CIFAR-10 trained on 16 GPUs with 16 samples given parallelization strategy and global batch size.}
\label{tab:mace_results}
\begin{tabular}{c c c c c}
\hline
Method/GBS & 256 & 512 & 1024 & 2048 \\
\hline
DDP & 0.1407 & 0.1399 & 0.1753 & 0.1525 \\
SP  & 0.1211 & 0.1121 & 0.0657 & 0.0534 \\
\hline
\end{tabular}
\end{table}

\subsection{MLP on ENTSO-E de}
\begin{figure}[ht]
    \centering
    \includegraphics[scale=0.4]{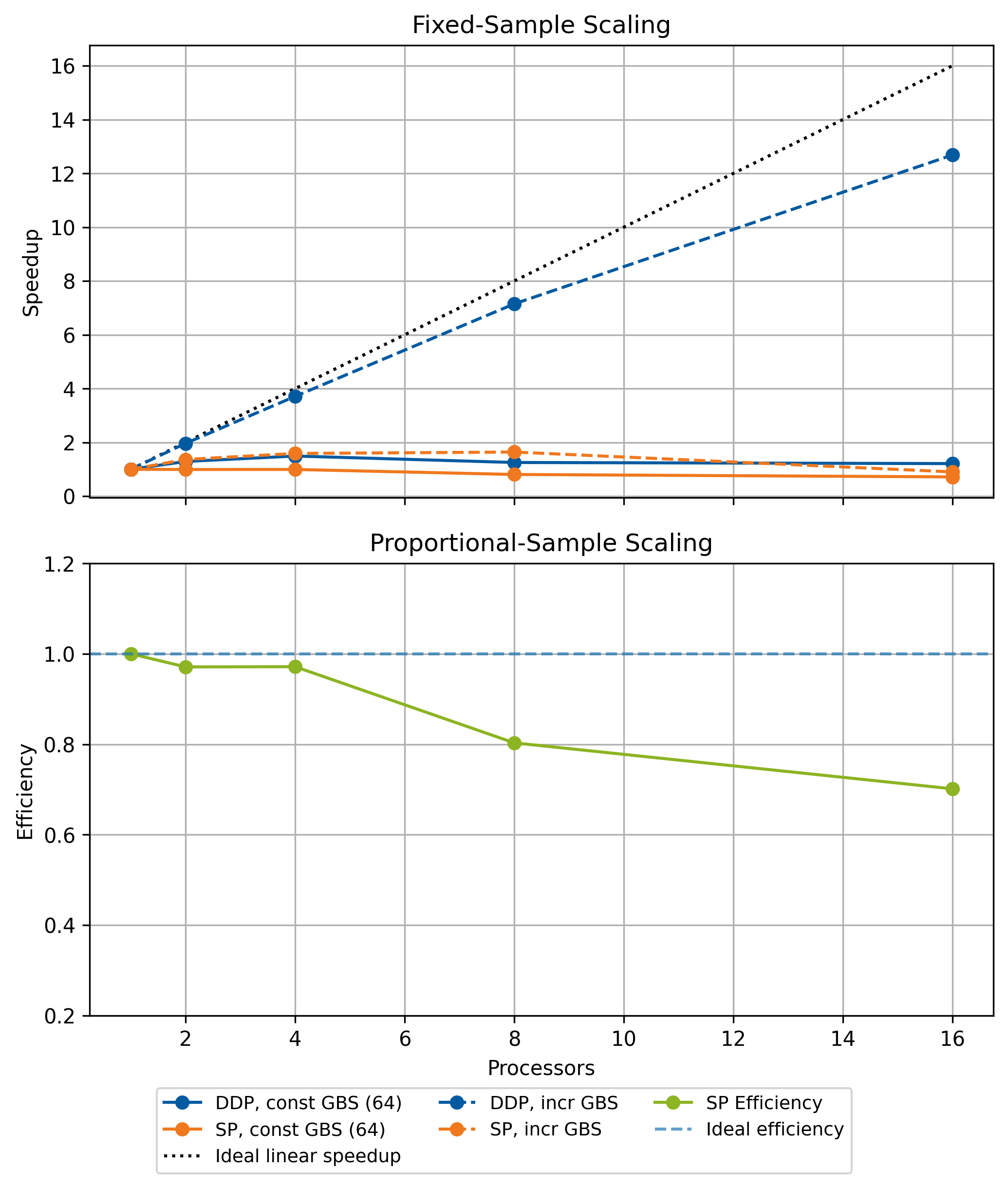}
    \caption{Speed up (fixed-sample scaling, top) and efficiency (proportional-sample scaling, bottom) of MCD on MLP with the ENTSO-E dataset. For fixed-sample scaling, we compare different sampling parallelism, data parallelism, and a hybrid of both, using 16 random samples and either a global batch size of 64 or an increasing global batch size of 64 times number of GPUs. For proportional-sample scaling we use a fixed global batch size of 1024 and scale the number of random samples from 1 to 16.}
    \label{fig:mlp-scaling}
    \Description{}
\end{figure}
%The local batch size was fixed to 64.
We further study the generalizability of the approach to different sampling-based Bayesian learning methods by transitioning from BNNs with VI to MCD, using the task of time series forecasting with an MLP. In a non-distributed setting, an epoch with 16 samples takes about 10 seconds for this task. 
Although sampling parallelism remains fully functional for the MLP in the sense that training can be carried out without instability, predictive performance matches expectations, and the communication patterns operate smoothly, its efficiency is considerably limited.

\Cref{fig:mlp-scaling} (bottom) shows the efficiency from proportional-sample scaling experiments. We observe that efficiency begins to decline sharply beyond four GPUs. This behavior is expected, as the communication overhead associated with sampling parallelism grows with the number of GPUs and quickly becomes significant compared to the relatively low computational cost of evaluating this small model.
In other words, for small networks, the cost of synchronizing gradients and managing multiple weight samples can outweigh the benefits of parallel computation. This highlights that the advantages of sampling parallelism are most pronounced for larger models, where the computational workload per GPU is substantial enough to amortize the communication overhead and achieve high scaling efficiency.

This effect becomes even more pronounced in fixed-sample scaling experiments (c.f.~\Cref{fig:mlp-scaling} top), where the workload per GPU decreases as more resources are added, eventually causing communication overhead to dominate and speedup to drop sharply. In contrast, DDP performs much better in these scenarios, as it parallelizes the data loading process, a step that cannot be fully vectorized on a single GPU, allowing for more effective utilization of additional GPUs.

\subsection{SWIN transformer on ERA5}
The previous experiments have demonstrated the general feasibility and efficiency of sampling parallelism, and the potential to even improve model convergence via data augmentation. 
However, the speed-up achieved through sampling parallelism is ultimately not competitive with DDP in those cases.
To demonstrate that by opening up an additional parallelization axis through sampling parallelism, which allows us to tackle the unprecedented challenges in large-scale Bayesian neural networks, we conduct proportional-sample scaling experiments on a SWIN Transformer for weather prediction using the ERA5 data. Performing fixed-sample scaling would require comparing results obtained on multiple GPUs to those from a single GPU running the same workload. Due to hardware limitations, it is not possible to accommodate the full workload on a single GPU. This limitation underscores the practical necessity of sampling parallelism for very large models. In a non-distributed setting, an epoch with 2 samples takes about 40 hours for this task. 

\begin{figure}[ht]
    \centering
    \includegraphics[scale=0.4]{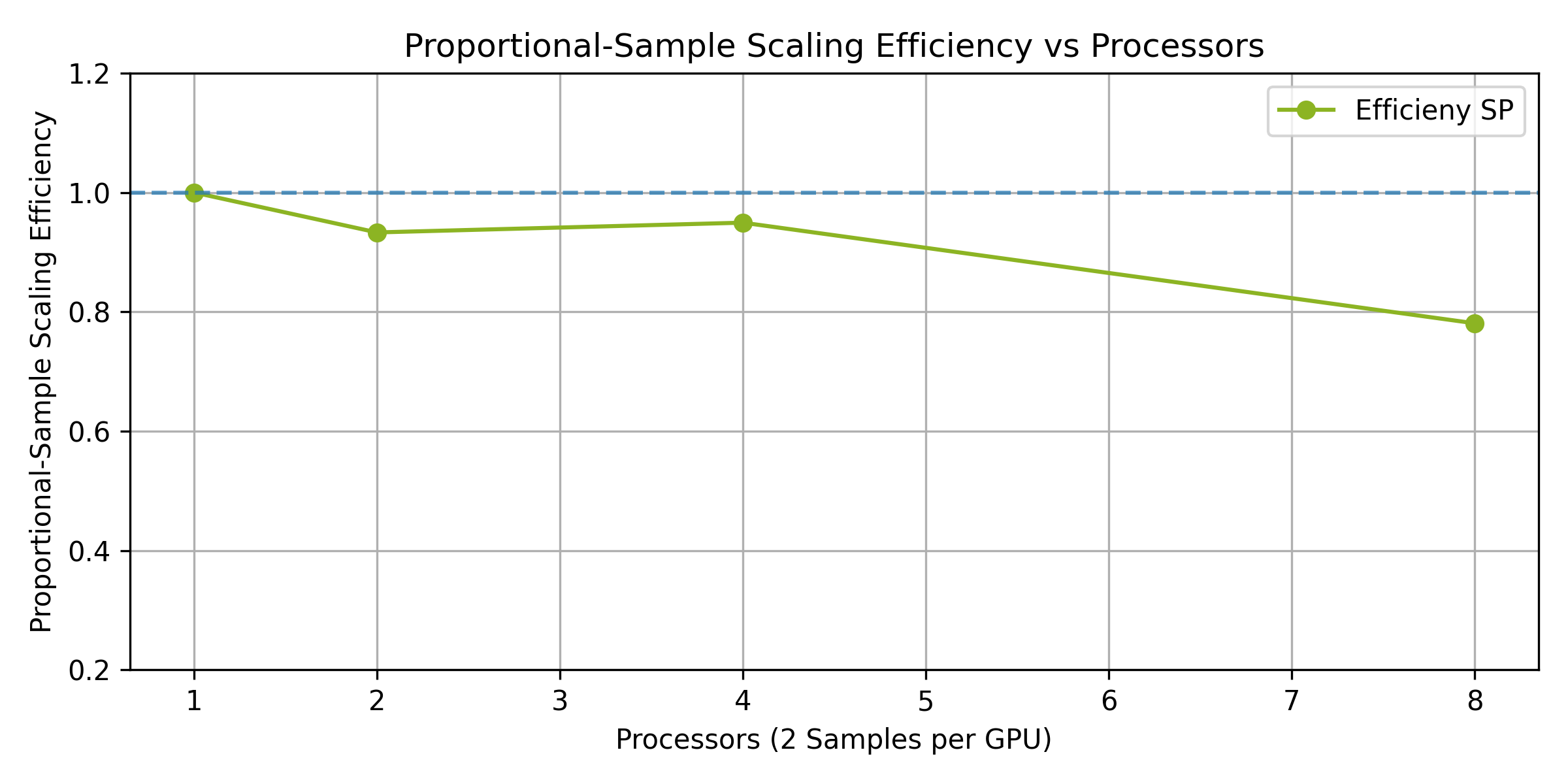}
    \caption{Scaling efficiency of the Bayesian SWIN Transformer with the ERA5 dataset, where the number of random samples equals 2 times the number of GPUs, and the local batch size per GPU is 1.}\label{efficiency_kangu}
    \Description{}
\end{figure}
Given that the used loss requires at least two random samples per GPU, the only configuration that fit into the GPU memory was a local batch size of one and exactly two random samples per GPU. Our proportional-sample scaling experiments depicted in~\Cref{efficiency_kangu} demonstrate the efficiency of scaling on up to 8 GPUs and 2 random samples per GPU, which enables training on 16 samples, which would not have been possible without parallelizing the samples. The proportional-sample scaling efficiency is at a reasonable level, remaining above $90\%$ for up to four GPUs and dropping to  $80\%$ for eight. Regardless of the scaling behavior, the fact that training on this many samples becomes possible shows, how important it is to be able to distribute these samples as this method allows training beyond the boundaries of DDP alone.

The presented results demonstrate that sampling parallelism is a valuable technique for sampling-based uncertainty quantification, particularly in settings where data augmentation plays a critical role in predictive performance.
%or where a seamless transition from a non-Bayesian to a Bayesian model is desired without extensive hyperparameter tuning.
Moreover, sampling parallelism can be flexibly integrated with other parallelization strategies, as illustrated by our hybrid approach, making it a versatile tool that can be tailored to the specific requirements of a given task. However, we do not advise to use sampling parallelism in very small networks, as the scalability is sub-par.

\section{Conclusion}

With this work, we aim to tackle the significant computational burden associated with training sampling-based methods, in particular BNNs, thereby facilitating their widespread practical adoption. Since this burden originates primarily from the need to evaluate multiple samples of the model parameter distributions, we introduce sampling parallelism as a strategy that distributes random sample evaluations across multiple processes. This approach makes large-scale Bayesian learning in neural networks substantially more tractable with respect to both memory requirements and runtime.
While duplicating data-loading operations across GPUs introduces overhead and can hinder raw speedup in comparison to other parallelization techniques such as DDP, it also enables each GPU to apply independent stochastic augmentations to the same batch. This increases the diversity of the training data, which can improve convergence and generalization. Thus, the method presents not only a computational trade-off but also a methodological opportunity. Another shortcoming of sampling parallelism is that each loss function that is used needs to be carefully examined to determine if simple gradient synchronization between GPUs is sufficient for exact replication of the sequential algorithm. If not, either further communication will be required or the parallelization will be an approximate replica whose effectiveness will need to be determined via experiments. As future work, developing a dedicated package that implements and documents loss functions known to preserve exactness under such parallelization schemes would be highly beneficial, as it would reduce the burden on practitioners and help standardize reliable usage. Nonetheless, our experiments show that the training behavior is not negatively affected in some use-cases.

Sampling parallelism complements existing parallelization techniques, such as data or model parallelism, by providing an additional axis of scalability. Our hybrid experiments illustrate that sampling parallelism can be combined effectively with other methods, allowing practitioners to tailor their parallelization strategy to the computational structure of their task.

Finally, sampling parallelism becomes especially valuable when both models and datasets grow large. 
 
When individual data samples or model components already saturate the memory of a single GPU (such as in atmospheric modeling), other parallelization techniques become a necessity.
In these cases, sampling parallelism provides a crucial additional degree of freedom for parallelizing and distributing a network, enabling uncertainty-aware learning.
While our experiments focus on BNNs and Monte Carlo Dropout, the underlying approach is applicable to a much broader family of Bayesian and UQ techniques. Evaluating performance across additional methods, architectures, tasks, and datasets will further map out its generality and practical scope, and allow us to position sampling parallelism as a practical and versatile tool for large-scale Bayesian modeling in real-world applications.

\begin{acks}
This work is supported by the German Federal Ministry of Research, Technology and Space (BMFTR) under the 01IS22068 EQUIPE grant and the 01LK2313A SMARTWEATHER21 grant, and by the Helmholtz AI platform grant. The authors gratefully acknowledge the computing time made available to them through the HAICORE@KIT partition and on high-performance computer HoreKa at the NHR Center KIT via the SmartWeather21-p0021348 NHR large project. This work was further supported by the Helmholtz Association's Initiative and Networking Fund on the HAICORE@FZJ partition.
\end{acks}

\bibliographystyle{ACM-Reference-Format}
\bibliography{sample-base}

\end{document}